%% file: main.tex
\documentclass{article}

\usepackage{arxiv}

\usepackage[utf8]{inputenc} 

\usepackage{amsfonts}       
\usepackage{nicefrac}       
\usepackage{microtype}      
\usepackage{lipsum}

\usepackage{color}
\usepackage{listings}
\usepackage{booktabs}
\usepackage{array}
\usepackage{graphicx}
\usepackage{longtable}
\usepackage{subfigure}

\usepackage{cite}
\usepackage{fancyhdr}

\usepackage{soul}

\usepackage{float}
\usepackage{listings}
\usepackage{mdwmath}
\usepackage{mdwtab}
\usepackage{multirow}
\usepackage{multicol}
\usepackage{setspace}
\usepackage[utf8]{inputenc}
\usepackage{lineno}
\usepackage{listings}

\usepackage{mdwmath}
\usepackage{mdwtab}
\usepackage{multirow}
\usepackage{multicol}
\usepackage{array}
\usepackage{booktabs}

\usepackage{enumitem}
\usepackage{xspace}
\usepackage[export]{adjustbox}
\usepackage{graphicx}
\usepackage{color,soul}
\usepackage{rotating}
\usepackage{setspace}
\usepackage{amsmath} 
\usepackage{amssymb}
\usepackage{float}
\usepackage{xcolor}

\usepackage{hyperref}
\usepackage[numbers]{natbib}

\usepackage{listings}
\usepackage[utf8]{inputenc}
\usepackage{supertabular}
\usepackage{mdwtab}
\usepackage{multirow}
\usepackage{multicol}
\usepackage{array}
\usepackage{booktabs}
\usepackage{longtable}
\usepackage{enumitem}
\usepackage{xspace}
\usepackage[export]{adjustbox}
\usepackage{graphicx}
\usepackage{mdframed}
\usepackage{soul}
\usepackage{makecell}
\usepackage{setspace}
\usepackage{float}
\usepackage{color, colortbl}
\usepackage{subfigure}
\usepackage{xcolor}
\usepackage{xurl}

\usepackage{balance}  

\usepackage[%
  para,
  online,
]{threeparttable}

\title{Responsible AI Pattern Catalogue: A Collection of Best Practices for AI Governance and Engineering}

\author{Qinghua Lu\textsuperscript{1}, Liming Zhu\textsuperscript{1}, Xiwei Xu\textsuperscript{1}, Jon Whittle\textsuperscript{1}, Didar Zowghi\textsuperscript{1}, Aurelie Jacquet\textsuperscript{1}\\
\textsuperscript{1}Data61, CSIRO, Australia\\
firstname.lastname@data61.csiro.au}

\begin{document}

\maketitle

\begin{abstract}
Responsible AI is widely considered as one of the greatest scientific challenges of our time and is key to increase the adoption of AI. Recently, a number of AI ethics principles frameworks have been published. However, without further guidance on best practices, practitioners are left with nothing much beyond truisms. Also, significant efforts have been placed at algorithm-level rather than system-level, mainly focusing on a subset of mathematics-amenable ethical principles, such as fairness. Nevertheless, ethical issues can arise at any step of the development lifecycle, cutting across many AI and non-AI components of systems beyond AI algorithms and models. To operationalize responsible AI from a system perspective, in this paper, we present a Responsible AI Pattern Catalogue based on the results of a Multivocal Literature Review (MLR). Rather than staying at the principle or algorithm level, we focus on patterns that AI system stakeholders can undertake in practice to ensure that the developed AI systems are responsible throughout the entire governance and engineering lifecycle. The Responsible AI Pattern Catalogue classifies the patterns into three groups: multi-level governance patterns, trustworthy process patterns, and responsible-AI-by-design product patterns. These patterns provide systematic and actionable guidance for stakeholders to implement responsible AI.

\end{abstract}

Responsible AI, ethical AI, trustworthy AI, AI governance, AI engineering, MLOps, software engineering, software architecture, pattern, best practice

\section{Introduction}
Artificial Intelligence (AI) has been transforming our society and listed as the top strategic technology in many organizations. Although AI has huge potential to solve real-world challenges, there are serious concerns about its ability to behave ethically and make decisions in a responsible way. 
Compared to traditional software systems, AI systems involve higher degree of uncertainty and more ethical risk due to their dynamic, autonomous and opaque decision making and historical-data-dependent behaviors. 
Responsible AI refers to the ethical development of AI systems to benefit the humans, society, and environment. The concept of responsible AI has attracted significant attention from governments, organizations, companies and societies.  
According to the 2022 Gartner CIO and Technology Executive Survey, 48\% of organizations have already adopted or plan to adopt AI technologies within the next 12 months while 21\% of organizations have already deployed or plan to deploy responsible AI technologies within the next 12 months\footnote{\url{https://www.gartner.com/en/articles/it-budgets-are-growing-here-s-where-the-money-s-going}}. Responsible AI has been widely considered as one of the greatest scientific challenges of our time and the key to unlock the market and increase the adoption of AI. 

To address the responsible AI challenges, a number of AI ethics principles frameworks have been published recently ~\cite{jobin2019global}, which AI systems are supposed to conform to. There has been a consensus made around the AI ethics principles~\cite{fjeld2020principled}. A principle-based approach allows technology-neutral, future-proof and context-specific interpretations and operationalization. However, without further best practice guidance, practitioners are left with nothing much beyond truisms. For example, it is a very challenging and complex task to operationalize the the human-centered value principle regarding how it can be designed for, implemented and monitored throughout the entire lifecycle of AI systems. In addition, significant efforts have been put on algorithm-level solutions which mainly focus on a subset of mathematics-amenable ethical principles (such as privacy and fairness). However, issues (including ethical issues) can occur at any step of the development lifecycle crosscutting many AI, non-AI and data components of systems beyond AI algorithms and models. To try to fill the principle-algorithmic gap, further guidance such as guidebooks\footnote{\url{https://www.microsoft.com/en-us/haxtoolkit/}}\footnote{\url{https://pair.withgoogle.com/guidebook}}, questions to generate discussions~\cite{liao2020questioning,liao2021question}, checklists~\cite{larasati2021ai,han2022checklist} and documentation templates~\cite{raji2020closing,jacovi2021formalizing,ahuja2020opening,hutchinson2021towards,zhou2022transparent,adkins2022prescriptive} have started to appear. Those efforts tend to be ad-hoc sets of more detailed prompts for practitioners to think about all the issues and come up with their own solutions.

\begin{figure}
\centering
\includegraphics[width=0.5\columnwidth]{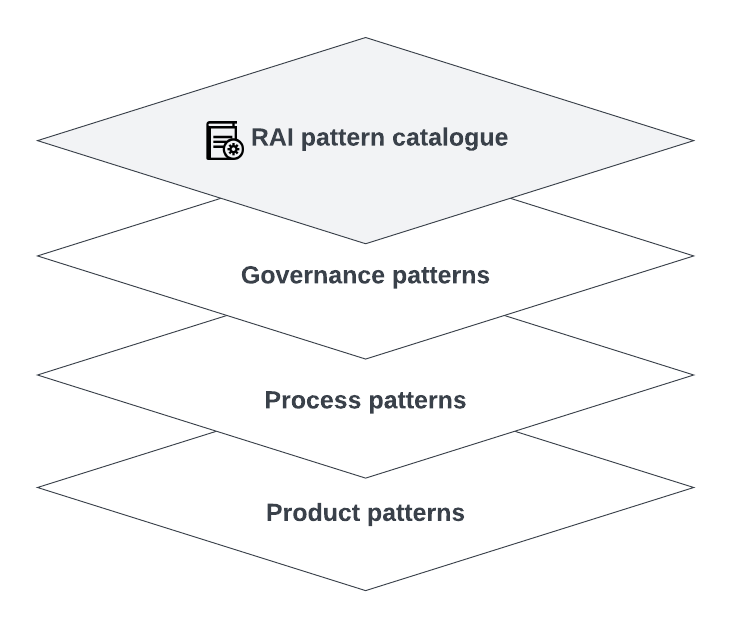}
\caption{Overview of RAI pattern catalogue.} \label{fig:overview}
\vspace{-2ex}
\end{figure}

In this paper, we therefore adopt a pattern-oriented approach and present a Responsible AI Pattern Catalogue for operationalizing responsible AI from a system perspective. In software engineering, a pattern is a reusable solution to a problem that occurs commonly within a given context in software development~\cite{Beck1987}. Rather than staying at the ethical principle level or algorithm level, we focus on patterns that practitioners can utilize in practice to ensure that the developed AI systems are responsible throughout the entire software development lifecycle. As shown in Fig.~\ref{fig:overview}, the Responsible AI Pattern Catalogue classifies patterns into three groups: 1) governance patterns for establishing multi-level governance for responsible AI; 2) process patterns for setting up trustworthy development processes; 3) product patterns for building responsible-AI-by-design paradigm into AI systems. These patterns are identified through conducting a systematic Multivocal Literature Review. The full version of our Responsible AI Pattern Catalogue can be accessed online \footnote{\label{catalogue}Responsible AI Pattern Catalogue: \url{https://research.csiro.au/ss/science/projects/responsible-ai-pattern-catalogue/} Each of the patterns is described following the traditional pattern structure (i.e., context, problem, solution, benefits, drawbacks,related patterns, known uses)}.

The remainder of the paper is organized as follows. Section 2 introduces the methodology for building up the pattern catalogue. Section 3 presents the AI system stakeholders and governance patterns. Section 4 discusses the process patterns for each stage of the development lifecycle. Section 5 introduces the project patterns. Section 6 discusses the related work. Section 5 concludes the paper.

\begin{figure}
\centering
\includegraphics[width=\columnwidth]{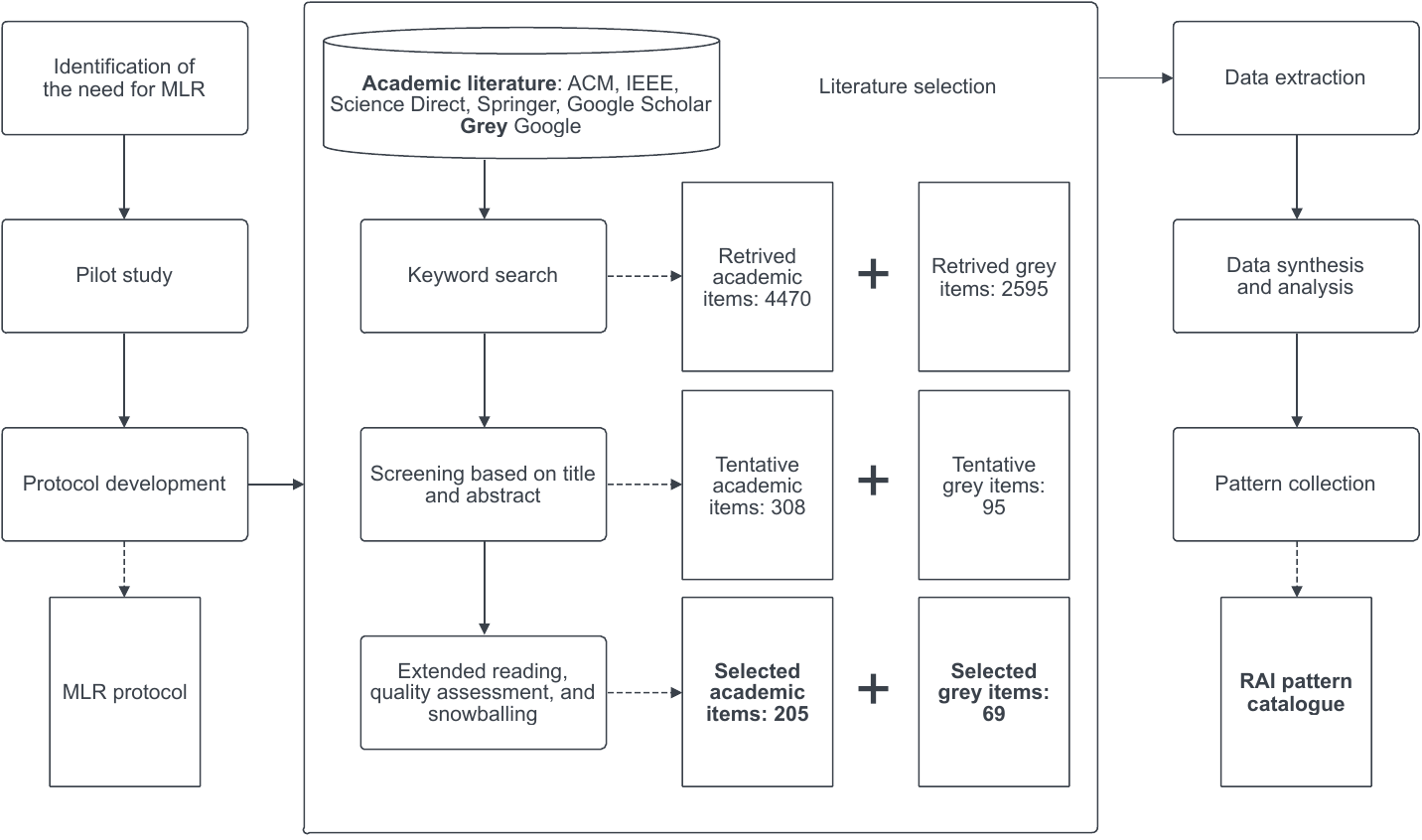}
\caption{Methodology.} \label{fig:methodology}
\vspace{-2ex}
\end{figure}

\section{Methodology}
To build up a Responsible AI (RAI) Pattern Catalogue, we performed a systematic Multivocal Literature Review (MLR) to collect patterns. Fig.~\ref{fig:methodology} presents the research design and methodology. The high level research question that has guided this research is: "What responsible AI solutions can be identified?". The research question focuses on identifying the reusable patterns for responsible AI.

\subsection{Data preparation}
The benefit of an MLR is to cover both academic literature and grey literature in the study. Grey literature is written by practitioners (such as governments, organizations, companies) and not published in books or scientific journals/conferences. However, grey literature can provide valuable insights on the state of practice and may include many industry solutions that are not discussed in academic papers. Given the nature of patterns, we decided to also review grey literature to understand the state of the practice in the field of Responsible AI and collect patterns from industry. In our MLR, we identify: (1) relevant academic peer-reviewed academic literature and (2) relevant grey literature for this study.

\subsection{Search strategy}
The study has been carried out separately for academic literature and grey literature. We adopted Kitchenham's Systematic Literature Review (SLR) guideline~\cite{kitchenham2007guidelines} to review the academic literature and used Garousi et al's guideline~\cite{GAROUSI2019101} to perform the grey literature review.
The complete MLR protocol is available as online material \footnote{\label{protocol}\url{https://drive.google.com/file/d/18Jiap714N1uprFVYU0jGmKSxKa-y2awJ/view?usp=sharing}}. Overall, we first tested different search strings in the five well-known search engines and evaluated the total number of studies retrieved, as well as their relevance. The evaluation involved cross-checking the inclusion of known relevant literature. Once we determined the most effective search string, we proceeded to perform searches on Google Scholar. From the results, we randomly selected 20 papers and extracted the relevant answers for each of the research questions.

We determined the search strings by deriving relevant keywords from the research question. Before conducting the systematic search, we did a pilot study by experimenting with the search terms to compare the results.
We used ``AI'', ``Responsible'', ``Solution'' as the key terms and included synonyms and abbreviations as supplementary terms to increase the search results. 
We designed the search strings for each primary source to check the title. 
After completing the first draft of search strings, we examined the results of each search string against each database to check the effectiveness of the search strings.
The finalised search terms are shown in Table~\ref{tab:key}.  We use Australia's AI ethics principles~\cite{DISER_2020} to identify the supplementary terms for ``Responsible'' as a close-enough representation of the many similar ones~\cite{jobin2019global,fjeld2020principled} around the world.  The eight AI ethics principles include \textit{human, societal and environmental wellbeing}, \textit{human-centred values}, \textit{fairness}, \textit{privacy protection and security}, \textit{reliability and safety}, \textit{transparency and explainability}, \textit{contestability}, \textit{accountability}. We mapped each individual term in the eight principles to its corresponding noun term and adjective forms. Furthermore, in order to encompass the relevant terms related to responsible AI, we expanded the mapping to include ``responsible'' as well as its variations, such as ethics, ethical, responsibility, trust, trusted, trustworthiness, and trustworthy. By doing so, we ensure comprehensive coverage of the terms relevant to responsible AI. The search strings and the respective paper quantities of the initial search for each primary source are listed in our MLR protocol.
We applied the search string to both scholar search engines for academic literature and Google Search Engines for grey literature. The scholar search engines include: ACM Digital Library, IEEE Xplore, Science Direct, Springer Link, and Google Scholar. The search period is until 31 July 2022. 

\begin{table*}[h]
  \caption{Key and supplementary search terms}
  \label{tab:key}
  \footnotesize
  \begin{tabular}{l|l}
    \toprule
    \textbf{Key Term} & \textbf{Supplementary Terms}\\
    \midrule
    \textup{{\makecell[l]{AI}}} & \textup{{\makecell[l]{Artificial Intelligence, Machine Learning, ML}}}\\
    \cmidrule(l){1-2}
    \textup{{\makecell[l]{Responsible}}} & \textup{{\makecell[l]{Ethics, Ethical, Responsibility, Trust, Trusted, Trustworthiness, Trustworthy, Human Values, Wellbeing,\\ Accountability, Accountable, Transparency, Transparent, Explainability, Explainable, Interpretability,\\ Interpretable, Contestability, Contestable, Fairness, Fair, Reliability, Reliable, Safety, Safe, Privacy, Private,\\ Security, Secure}}}\\
    \cmidrule(l){1-2}
    \textup{{\makecell[l]{Solution}}} & \textup{{\makecell[l]{Tactic, Practice, Process, Design, Architecture, Solution, Approach, Method, Mechanism, Tool, Toolkit}}}\\
    \bottomrule
  \end{tabular}
\end{table*}

We screened the initial results against inclusion and exclusion criteria. The inclusion criteria include: (1) A paper/article that presents a governance or process or design solution for responsible AI.
(2) A paper/article that presents a tool or toolkit for developing responsible AI systems. (3) A paper/article that is in the form of a published scientific paper or industry article.
The exclusion criteria are: (1) A paper/article that only discusses high-level principles or frameworks. (2) A paper/article that only focuses on algorithm-level techniques. (3) A paper/article that is not written in English. (4) Conference version of a study that has an extended journal version. (5) PhD/Master’s dissertations, tutorials, editorials, books. 

The snowballing technique has been recommended and used in place of database searches in systematic literature reviews. For the academic literature, we identified a set of papers that serve as the starting point (i.e., seed set) for snowballing. The seed set papers were selected based on the source databases and Australia’s AI ethics principles to cover various communities. For each of the 5 source databases, we selected 1 top cited paper for responsible AI in general and each of the 8 principles respectively. For some principles, there is no paper found in one particular database. 
We only collected the grey literature from the first 10 Google pages. 
For the grey literature, snowballing is conducted if related responsible AI solutions are mentioned on the webpage.
We finally identified 205 academic items and 69 grey items for the MLR. For the grey literature, we organized the responsible AI solutions according to the corresponding companies. For example, we found 13 responsible AI tools/solutions on Microsoft's website but only counted Microsoft as one grey item in our data extraction sheet and recorded a few patterns extracted from Microsoft's tools/solutions.


\subsection{Data extraction, synthesis, and analysis}

To realise responsible AI from a software engineering perspective, we need to make both AI products and their development processes trustworthy and responsible. Additionally, compliance with AI standards and laws from a governance perspective is necessary.  Thus, we classify the patterns into three categories: governance, process, and product. Not only should you use product patterns to enforce responsible AI principles directly in the product and verify/validate the product, but you should also use process and governance patterns to complement it further.

We extracted data and summarized findings from the selected academic and grey items based on the pre-defined research question. Based on the answers extracted for the research question, we identified different types of patterns. For example, there are a few papers using federated learning to deal with data privacy issues, thus ``federated learner'' is identified as a product pattern that can be built into the architecture of AI systems for continuous learning. Some of the solutions can be mapped to multiple levels. For example, software bill of materials can be identified as an organisation-level governance pattern which is interconnected with and supported by product pattern ``Bill of materials registry''. After identifying a pattern, we documented its details on our RAI Pattern Catalogue website according to the traditional pattern structure~\cite{martin1997pattern}: context, problem, solution, benefits, drawbacks, related patterns, known uses. The known uses were found through data extraction and additional manual search. The pattern users and impacted stakeholders are identified based on 1) AI software supply chain and ecosystem, 2) AI standards, 3) our expertise and knowledge.


We also extract some general information, e.g. authors name, organization, publication venue, publication year. We performed a pilot study on 20 items to test the research question and the way to extract the required data. We stored all the extracted data in a spreadsheet for analysis.

\begin{figure}
\centering
\includegraphics[width=\columnwidth]{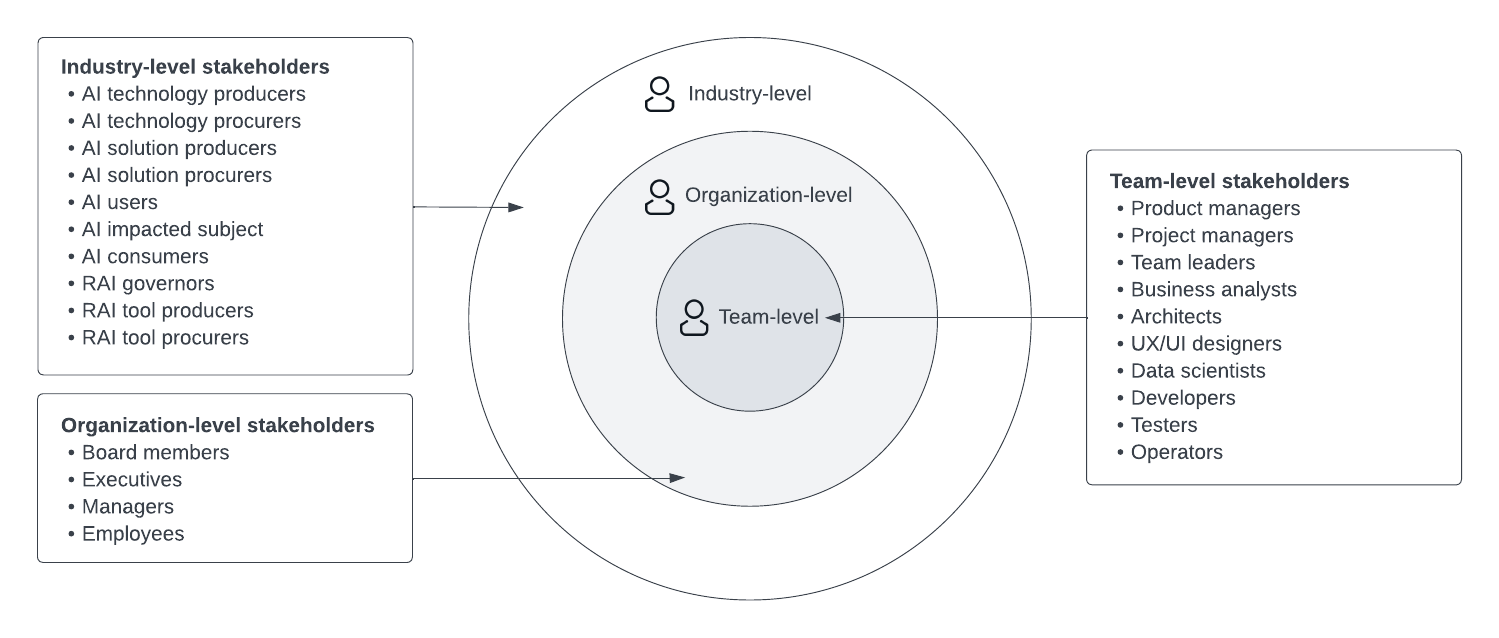}
\caption{Stakeholders for RAI governance.} \label{fig:stakeholders}
\vspace{-2ex}
\end{figure}

\section{Governance Patterns}
The governance for responsible AI systems can be defined as the structures and processes that are employed to ensure the development and use of AI systems meet AI ethics principles. According to Shneiderman’s structure~\cite{Shneiderman20}, governance can be built at three levels: industry-level, organization-level, and team-level. 
As illustrated in Fig.~\ref{fig:stakeholders}, we identified the stakeholders for RAI governance and classified them into three groups: 

\begin{itemize}
    \item \textbf{Industry-level stakeholders}
    
    \begin{itemize}
        \item \textbf{AI technology producers}: develop AI technologies for others to build on top to produce AI solutions, e.g., parts of Google, Microsoft, IBM. AI technology producers may embed RAI in their technologies and/or provide additional RAI tools.
        
        \item \textbf{AI technology procurers}: procure AI technologies to build their in-house AI solutions, e.g., companies or government agencies buying/using AI platform/tools. AI technology procurers may care about RAI issues and embed RAI into their AI technology procurement process.
        
        \item \textbf{AI solution producers}: develop in-house/blended unique solutions on top of technology solutions and need to make sure the solutions adhere to RAI principles/standards/regulations, e.g., parts of MS/Google providing Office/Gmail ``solutions''. They may offer the solutions to AI consumers directly or sell to others.  They may use RAI tools (provided by AI technology producers or RAI tool producers) and RAI processes during their solution development.
        
        \item \textbf{AI solution procurers}: procure complete AI solutions (with some further configuration and instantiation) to use internally or offer to external AI consumers, e.g., a government agency buying from a complete solution from vendors. They may care about RAI issues and embed RAI into their AI solution procurement process.
        
        \item \textbf{AI users}: who use an AI solution to make decisions that may impact on a subject, e.g., a loan officer or a gov employee. AI users may exercise additional RAI oversight as the human-in-the-loop.
        
        \item \textbf{AI impacted subjects}: who are impacted by some AI-human dyad decisions, e.g., a loan applicant or a tax payer. AI impacted subjects may care about RAI issues and contest the decision on dyad AI ground. 
        
        \item \textbf{AI consumers}: who consume AI solutions (e.g., voice assistants, search engines, recommender engines) for their personal use (not affecting 3rd parties). AI consumers may care about RAI issues and the dyad AI aspects of AI solutions.
        
        \item \textbf{RAI governors}: those that set and enable RAI policies and controls within their culture. RAI governors could be functions within an organization in the above list or external (regulators, consumer advocacy groups, community).
        
        \item \textbf{RAI tool producers}: technology vendors and dedicated companies offering RAI features integrated into AI platforms or AIOps/MLOps tools.
        
        \item \textbf{RAI tool procurers}: any of the above stakeholders who may purchase or use RAI tools to improve or check solutions/technology’s RAI aspects.
        
    \end{itemize}
    
    \item \textbf{Organization-level stakeholders}
    
    \begin{itemize}
        \item \textbf{Management teams}: individuals at the higher level of an organization who are responsible for establishing RAI governance structure in the organization and achieving RAI at the organization-level. The management teams include board members, executives, and (middle-level) managers for legal, compliance, privacy, security, risk, and sustainability.
         \item \textbf{Employees}: individuals who are hired by an organization to perform work for the organization and expected to adhere to RAI principles in their work.
    \end{itemize}
    
    \item \textbf{Team-level stakeholders}
    
    \begin{itemize}
        \item \textbf{Development teams}: those who are responsible for developing and deploying AI systems, including product managers, project managers, team leaders, business analysts, architects, UX/UI designers, data scientists, developers, testers, and operators. The development teams are expected to implement RAI in their development process and embed RAI into the product design of AI systems.

        
        
        
        
        
        
        
        
        
    \end{itemize}
\end{itemize}

\begin{figure}
\centering
\includegraphics[width=\columnwidth]{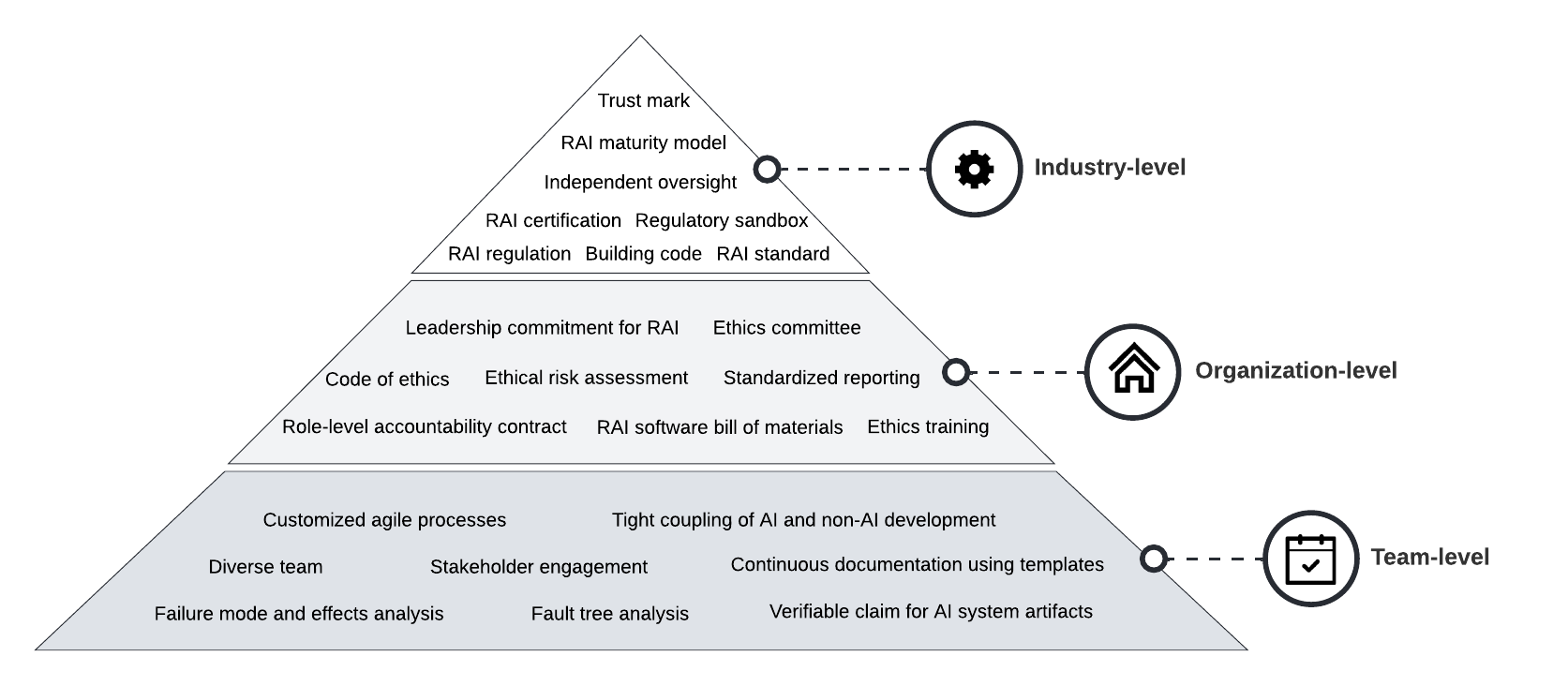}
\caption{Governance patterns for responsible AI.} \label{fig:governance}
\vspace{-2ex}
\end{figure}

As shown in Fig.~\ref{fig:governance}, we identify a set of governance patterns and classify them into industry-level governance patterns, organization-level governance patterns, and team-level governance patterns based on Shneiderman's governance structure~\cite{Shneiderman20}. The target users of industry-level governance patterns are RAI governors, while the impacted stakeholders include AI technology producers and procurers, AI solution producers and procurers, RAI tool producers and procurers. For the organization-level patterns, the target users are the management teams and the impacted stakeholders are employees, AI users, AI consumers, and AI impacted subjects. The target users of team-level patterns are the development team, whilst the impacted stakeholders are AI users, AI consumers, and AI impacted subjects.

\subsection{Industry-level governance patterns}

\subsubsection{RAI regulation}\hfill\\
Laws already apply to AI systems, however the processes/requirements to ensure compliance are not always certain, also some regulations may need to be updated e.g. administrative law. There is an urgent need for clear guidance to ensure that AI systems are developed and used responsibly in compliance with existing and upcoming laws, e.g., discrimination law. RAI regulations are developed by governments in their jurisdiction to enable the trustworthy development of AI systems by industry ~\cite{Shneiderman20, shneiderman2021responsible,jackson2021ethics,schaich2021four, papagiannidis2021deploying, ibanez2021operationalising, dignum2019ensuring}. Organisations will be required to ensure that they comply with the requirements of the EU AI Act when the applications fall into the high risk category~\footnote{\url{https://artificialintelligenceact.eu}}. In US, the Algorithmic Accountability Act of 2022~\footnote{\url{https://www.congress.gov/bill/117th-congress/house-bill/6580/text?r=2&s=1}} was introduced in the Senate and House of Representatives, while an AI Bill of Rights~\footnote{\url{https://www.wired.com/story/opinion-bill-of-rights-artificial-intelligence/}} is under development by the White House Office of Science and Technology Policy. The aim of RAI regulations is to prevent illegal or negligent, malicious use of AI systems. However, there are many regulations in developments in each jurisdictions, which may cause an interoperability challenge for organisations. Also, it usually takes a long time to enact AI regulations due to the lengthy consultation and approval process.

\subsubsection{Regulatory sandbox}\hfill\\
To enable the trial of the innovative AI products in the market, a regulatory sandbox can be designed to allow testing the innovative AI products in the real-world under relaxed regulatory requirements but with appropriate safeguards in place on a time-limited and small-scale basis~\cite{schaich2021four}. An AI Regulatory Sandbox \footnote{\url{https://www.eipa.eu/publications/briefing/sandboxes-for-responsible-artificial-intelligence/}} is introduced in the EU’s AI Act proposal submitted in 2021. The UK Information Commissioner’s Office advised a Regulatory Sandbox \footnote{\url{https://ico.org.uk/for-organisations/regulatory-sandbox/the-guide-to-the-sandbox/}} for utilising personal data. The Australian Government released the Enhanced Regulatory Sandbox\footnote{\url{https://asic.gov.au/for-business/innovation-hub/enhanced-regulatory-sandbox/}} for innovative financial services. 
AI products can enter the market under more flexible regulatory requirements in a faster pace and be tested in the real-world market to ensure they are designed ethically. However, it might incur extra cost to apply for a regulatory sandbox. Also, the AI products might not work well with large scale deployment in different context.

\subsubsection{Building code}\hfill\\
AI systems may have various degrees of risk depending on the design and application domains.
To ensure that AI systems are trustworthy and meet certain minimum standards, 
building code can be designed to provide mandatory regulatory rules for authority parties (e.g., independent oversight and advisory committee) to assess the compliance of AI systems before they are allowed to launch~\cite{Shneiderman20}. For example, IEEE has released a set of building codes for developing smart cities~\footnote{\url{https://cybersecurity.ieee.org/blog/2017/10/04/building-code-for-the-internet-of-things/}}, Medical Device Software Security ~\footnote{\url{https://ieeecs-media.computer.org/media/technical-activities/CYBSI/docs/BCMDSS.pdf}}, and Power System Software Security ~\footnote{\url{https://ieeecs-media.computer.org/media/technical-activities/CYBSI/docs/BCPSSS.pdf}}. Building code sets out clear compulsory regulatory requirements for developing AI systems. AI systems cannot be sold in the market until an approval is issued by the assessment authority.

\subsubsection{RAI standard}\hfill\\
An AI system may use data or components from multiple jurisdictions where may have conflicting regulatory requirements on their usage. 
To enable interoperability between jurisdictions, RAI standards are developed to describe repeatable processes to develop and use AI systems responsibly that are recognised internationally and can be either mandated by law or by contract~\cite{Shneiderman20,shneiderman2021responsible,ibanez2021operationalising}. ISO/IEC JTC 1/SC42 AI Technical Committee is developing ISO/IEC 42001 IT-AI-Management System Standard~\footnote{\url{https://www.iso.org/standard/77304.html}} that provides a pathway for the certification of AI systems and WG3 trustworthiness that covers risk management and bias~\footnote{\url{https://www.iso.org/committee/6794475.html}}. IEEE has released Guide for Architectural Framework and Application of Federated Learning~\footnote{\url{https://standards.ieee.org/ieee/3652.1/7453/}}, Standard for Technical Framework and Requirements of Trusted Execution Environment based Shared Machine Learning~\footnote{\url{https://standards.ieee.org/ieee/2830/10231/}}, and IEEE p7000 IEEE Standards for Model Process for Addressing Ethical Concerns During System Design ~\footnote{\url{https://ethicsinaction.ieee.org/p7000/}}.
Those AI standards provide repeatable processes and guidance for the use and development of AI systems that are recognised internationally. 

\subsubsection{RAI maturity model}\hfill\\
Organizations can face challenges that can hurt their business if they are not aware of their RAI maturity. RAI maturity model can be used to assess an organization’s RAI capabilities and the degree of readiness to take advantage of AI based on a set of dimensions~\cite{Shneiderman20,zhobe2021magic,fukas2021developing, alsheiabni2019towards}. RAI maturity model can guide organizations on how to increase their RAI capabilities. The assessment results depend on the model quality, e.g., assessment dimensions and rating methods. There have been a few AI maturity models developed in industry, such as Gartner’s AI Maturity Model~\footnote{\url{https://www.gartner.com/smarterwithgartner/the-cios-guide-to-artificial-intelligence}},  Microsoft’s AI Maturity Model~\footnote{\url{https://query.prod.cms.rt.microsoft.com/cms/api/am/binary/RE4DIvg}}, and IBM’s AI Maturity Framework~\footnote{\url{https://www.ibm.com/downloads/cas/OB8M18WR}}.  

\subsubsection{RAI certification}
\hfill\\ 
AI is a high-stake technology that requires evidence to prove AI products' compliance with AI standards or regulations in order to operate in the society. 
RAI certification can be designed to recognize that an organization or a person has the ability to develop or use an AI system in a  way that is compliant with standards or regulations~\cite{Shneiderman20, shneiderman2021responsible,henriksen2021situated,martin2017taking,yap2020towards,dignum2019ensuring,cihon2021ai,boza2021implementing,luxton2014recommendations}. Malta AI-ITA certification~\footnote{\url{https://mdia.gov.mt/wp-content/uploads/2019/10/AI-ITA-Guidelines-03OCT19.pdf}} is the world’s first AI certification scheme for RAI systems. 
DO-178C Certification~\footnote{\url{https://my.rtca.org/nc__store?search=do-178c}} has been used to approve commercial software-based aerospace systems. Queen’s University offers an executive education program on Principles of AI Implementation~\footnote{\url{https://smith.queensu.ca/ConversionDocs/Execdev/Trusted_Data_AI_Canadian_Business.pdf}}. 
The evidence of compliance can be provided through RAI certification to improve human trust in AI systems. However, like other types of certificates, RAI certificates may be forged, which makes the verification of authenticity of certificates challenging. The certification process is usually complex, costly and time consuming.

\subsubsection{Trust mark}
\hfill\\
Consumers in the market usually do not have professional knowledge about AI. 
To improve public confidence on AI and dispel their ethical concerns, trust mark, a seal of endorsement, is easy to understand by all consumers and can be used to inform consumers about AI system. Trust mark is important for small companies which are often not well-known in the AI market. However, consumers may not trust the AI systems with trust mark performing more responsible than those without one. There have been several trust marks designed for responsible use of data, such as Australian Data and Insights Association Trust Mark~\footnote{\url{https://dataandinsights.com.au/trust-mark/}},  New Zealand Privacy Commissioner's Privacy Trust Mark~\footnote{\url{https://www.privacy.org.nz/resources-2/applying-for-a-privacy-trust-mark/}}, and Singapore's Data Protection Trustmark (DPTM)~\footnote{\url{https://www.pdpc.gov.sg/overview-of-pdpa/data-protection/business-owner/data-protection-trustmark}}.  

\subsubsection{Independent oversight}
\hfill\\ 
Decisions made by AI systems may lead to severe failures due to its autonomous decision-making process. To audit AI systems and investigate failures in a trusted way, independent oversight can be conducted by the independent oversight boards that consist of experts who are knowledgeable to perform the review and has no conflict of interest with the reviewed organizations\cite{Shneiderman20,shneiderman2021responsible,mccradden2020patient,zhu2022ai}. 
 The U.S. National Transportation Safety Board~\footnote{\url{https://www.ntsb.gov/}} investigates every civil aviation accident.
The U.S. National Artificial Intelligence Advisory Committee ~\footnote{\url{https://www.ai.gov/naiac/\#ABOUT-NAIAC}} advises on AI-related issues.
Independent oversight provides a trusted review infrastructure to gain public confidence.
Planning oversight provides early feedback on the new development proposals. Failures of independent oversight could happen due to lack of sufficient independence.

\subsection{Organization-level governance patterns}
\subsubsection{Leadership commitment for RAI}
\hfill\\
The management teams need to understand the values, cost and risk for adopting AI in an the organization. Commitment needs to be made by the management team to build RAI culture within an organization~\cite{Shneiderman20}. Leadership commitment is achieved by the management team dedicating their time and efforts on establishing ethics principles and governance structure (e.g., appointment of chief RAI officer, RAI advisory boards)~\cite{sloane2022german}, as well as incorporating RAI into organization's values, vision, mission~\cite{shneiderman2021responsible}, board's strategy planning, executives' performance reviews\cite{schaich2021four}, audit and risk committee's scope~\cite{jackson2021ethics}, and ESG commitments. Leadership commitment enables organizational culture on RAI and visible sponsorship to build RAI capability. IBM has established an AI ethics board~\footnote{\url{https://www.ibm.com/au-en/artificial-intelligence/ethics}} to support a culture of RAI throughout IBM. Axon has assembled an independent AI ethics board~\footnote{\url{https://www.axon.com/company/ai-and-policing-technology-ethics}} to provide guidance on AI system development. Schneider Electric has appointed its first Chief AI Officer ~\footnote{\url{  https://www.iteuropa.com/news/schneider-electric-appoints-new-caio-and-opens-new-ai-hub}} to advance its AI strategy.

\subsubsection{Ethics committee}
\hfill\\
Organizations need to build capability incorporating multiple areas of expertise to address RAI issues. An AI ethics committee is an AI governance body that is established to develop standard processes for decision-making, as well as to approve and monitor AI projects~\cite{martin2017taking,boza2021implementing}.
Accenture released a report on how to build AI ethics committees~\footnote{\url{https://www.accenture.com/au-en/insights/software-platforms/building-data-ai-ethics-committees}}. Adobe has created an AI ethics committee~\footnote{\url{https://www.adobe.com/about-adobe/aiethics.html}} which includes experts with different background. Sony has established AI ethics committee~\footnote{\url{https://www.sony.com/en/SonyInfo/sony_ai/responsible_ai.html}} to ensure the ethically development of AI systems. The ethics committee provides feedback and guidance to the project team after reviewing the proposal. However, the committee might not have the expertise to review a particular case, which might cause bias issues.

\subsubsection{Code of ethics for RAI}
\hfill\\
AI may make wrong decisions or behave inappropriately, e.g. impact human lives or buy the wrong product. To guide AI related activities in an organization, a code of ethics is a set of rules that employees should uphold when developing an AI system~\cite{Shneiderman20,martin2017taking,hooker2018toward,ibanez2021operationalising}. 
AAAI has issued Code of Professional Ethics and Conduct~\footnote{\url{https://www.aaai.org/Conferences/code-of-ethics-and-conduct.php}} for all members.
Bosch sets out code of ethics~\footnote{\url{https://www.bosch-ai.com/industrial-ai/code-of-ethics-for-ai/}} to establish guidelines for the development of AI.
BMW has released a code of ethics for AI ~\footnote{\url{https://www.press.bmwgroup.com/global/article/detail/T0318411EN/seven-principles-for-ai-bmw-group-sets-out-code-of-ethics-for-the-use-of-artificial-intelligence}}.
A code of ethics provides employees with the same concrete rules on developing AI systems, but it relies on individuals to do the right thing with limited monitoring and enforcement.

\subsubsection{Ethical risk assessment}
\hfill\\
Although there are increasing concerns on AI ethics, RAI regulation is still at a very early stage. To assess the ethical risks associated with AI systems, an organization needs to extend the existing IT risk framework or design a new one to cover AI ethics~\cite{Shneiderman20,martin2017taking,d2018towards,jackson2021ethics,benjamins2019responsible,eitel2021beyond,raji2020closing,ahuja2020opening,lee2021risk}.
ISO/IEC JTC 1/SC 42 committee is developing ISO/IEC 23894 on Artificial Intelligence and Risk Management~\footnote{\url{https://www.iso.org/standard/77304.html}}.
NIST released the initial draft of AI Risk Management Framework that provides a standard process for managing risks of AI systems~\footnote{\url{https://www.nist.gov/itl/ai-risk-management-framework}}.
The Canadian government has released the Algorithmic Impact Assessment tool  to identify the risks associated with automated decision-making systems~\footnote{\url{https://www.canada.ca/en/government/system/digital-government/digital-government-innovations/responsible-use-ai/algorithmic-impact-assessment.html}}.
The Australian NSW government is mandating all its agencies that are developing AI systems to go through the NSW AI Assurance Framework~\footnote{\url{https://www.digital.nsw.gov.au/policy/artificial-intelligence/nsw-ai-assurance-framework}}. Singapore launches AI Verify Toolkit to test responsible AI~\footnote{\url{https://file.go.gov.sg/aiverify.pdf}}. UK ICO released ai and data protection risk toolkit~\footnote{\url{https://ico.org.uk/for-organisations/guide-to-data-protection/key-dp-themes/guidance-on-ai-and-data-protection/ai-and-data-protection-risk-toolkit/}} which is built up on their guidance for organizations using AI systems. Although ethical risk assessment has the potential to prevent majority of incidents and increase awareness of RAI, it is often a one-off type of risk assessment with subjective judgement on measurement~\cite{redmill2002risk}. 

\subsubsection{Standardized reporting}
\hfill\\
Standardized reporting is essential to address the opaque black box issue of AI systems. Organizations should set up standarized processes and templates for informing the development process and product design of AI systems to different stakeholders (e.g., AI governors, users, consumers)~\cite{schultz2022towards}. RAI regulations may request such obligations to ensure transparency and explainability of AI systems. The Cyberspace Administration of China published transparent disclosure requirements for online service providers~\footnote{\url{http://www.cac.gov.cn/2022-01/04/c_1642894606364259.htm}}. The service providers are requested to file with the regulators (i.e., AI governors) for impact assessment when realising new services. In addition, the online services must inform users when AI is being used to recommend contents to them and explain the purposes and design of recommended systems.
In EU's AI Act~\footnote{\url{https://eur-lex.europa.eu/legal-content/EN/TXT/?uri=CELEX\%3A52021PC0206}}, the incidents of AI systems are required to be reported and disclosed by AI system providers (i.e., AI technology or solution producers). Helsinki~\footnote{\url{https://ai.hel.fi/en/ai-register/}} and Amsterdam~\footnote{\url{https://algoritmeregister.amsterdam.nl/en/ai-register/}} released AI registers describing where and how the two cities are using AI, how AI is built, which data and algorithms are used, how the applications impact the citizens' daily lives, and development team's contact information.

\subsubsection{Role-level accountability contract}
\hfill\\
It is necessary that organizations have an appropriate approach to enable accountability throughout the entire lifecycle of AI systems. Role-level accountability can be established through formal contracts to define the boundary of responsibility and identify who should be held accountable when an AI system misbehaves~\cite{zhu2022ai}. For example, Australia’s National Data Commissioner creates a data sharing agreement template for using Australian Government data~\footnote{\url{https://www.datacommissioner.gov.au/data-management/data-sharing-agreement}}. Developers primarily focus on the technique aspects of AI systems and may not be familiar with the ethical principles. Role-level accountability contracts make the developers keep ethics in mind at every step, but may create stress for employees at all levels within an organization.

\subsubsection{RAI software bill of materials}
\hfill\\
From a software supply chain angle, development of AI systems involves complex and dynamic software supply chain. Many organizations procure AI technologies/solutions to build their AI systems. The AI systems are often assembled by using commercial or open-source AI and/or non-AI components from third parties. Despite cost efficiency, the underlying security and integrity issues of the third party components have attracted significant attentions. According to Sonatype's report on 2021 state of the software supply chain~\footnote{\url{https://www.sonatype.com/resources/state-of-the-software-supply-chain-2021}}, the software supply chain attacks increased 650\% in 2021, while it was 430\% in 2020. RAI software bill of materials keeps a list of components used to create an AI software product, which can be used by AI solution procurers and consumers to check the supply chain details of each component of interest and make buying decisions~\cite{barclay2019towards}. 
The supply chain details should at least include component name, version, supplier, dependency relationship, author of software bill of materials data, and timestamp~\cite{sbom_mini}.
This provides traceability and transparency about components and allows AI solution procurers and consumers to easily check component information (such as supply chain details and context information) and track ethical issues. RAI software bill of materials enables faster vulnerability identification but may need to be updated frequently since AI systems may evolve over time. Dependency-Track~\footnote{\url{https://dependencytrack.org/}} is widely used by practitioners to track components’ supply chain information and identify known vulnerabilities. Software Package Data Exchange (SPDX)~\footnote{\url{https://spdx.dev/}} and CycloneDX~\footnote{\url{https://cyclonedx.org/}} are two standards for exchanging software bill of material information for security analysis.

\subsubsection{Ethics training}
\hfill\\
It is urgent that the employees of an organizations begin to think through the potential implications of AI on their work and make ethical choices during the development and use of AI systems. Ethics training provides employees with knowledge on how to deal with ethical issues during development~\cite{Shneiderman20,kasenberg2018norms,shneiderman2021responsible,schaich2021four,sand2022responsibility,eitel2021beyond,ibanez2021operationalising,zhu2022ai,amershi2019software}. MIT offers a 3-day course "Ethics of AI: Safeguarding Humanity"~\footnote{\url{https://professional.mit.edu/course-catalog/ethics-ai-safeguarding-humanity}} introducing the ethics of AI development and deployment. The University of Technology Sydney (UTS) designed a short course "Ethical AI: from Principles to Practice"~\footnote{\url{https://open.uts.edu.au/uts-open/study-area/Technology/ethical-ai-from-principles-to-practice/}} for business executives. The University of Helsinki created a free online course "The Ethics of AI"~\footnote{\url{https://ethics-of-ai.mooc.fi/}} for anyone who is interested in AI ethics. Halmstad University provides a short course on critical design and practical ethics for AI~\footnote{\url{https://learning4professionals.se/showCourse/158/Critical_Design_and_Practical_Ethics_for_AI}}. Ethics training can improve organizational awareness on RAI and sharpen the employees’ RAI skills. However, RAI covers a broad range of knowledge and skills and ethics training may only offer a subset of knowledge and skills within limited time.

\subsection{Team-level governance patterns}
\subsubsection{Customized agile process}
\hfill\\
Agile development has been increasingly adopted by organizations to incrementally and iteratively develop software systems, including AI systems. However, the existing agile development methods mainly focus on business value and largely neglect the AI ethics principles. 
To address ethical issues in the AI system development process, agile methods need to be extended and customized to allow consideration of ethics principles. Extension points could be artefacts, roles, ceremonies, practices, and culture~\cite{hussain2022can}. Microsoft’s Azure DevOps allows the customization of inherited processes\footnote{\url{https://docs.microsoft.com/en-us/azure/devops/organizations/settings/work/inheritance-process-model?view=azure-devops&tabs=agile-process}}. Atola Technology provides customized agile methodology that contains different development practices\footnote{\url{https://www.airtable.com/universe/exp4OppRObzXbhOQE/custom-agile-methodology-by-atola}}. Apptio Targetprocess is a web-based visual tool for managing projects with flexibility at various levels\footnote{\url{https://www.apptio.com/products/targetprocess/}}.

\subsubsection{Tight coupling of AI and non-AI development}
\hfill\\
AI system development involves the development of both AI and non-AI components with rapid iterations. This requires more frequent integration of AI and non-AI components. Compared with non-AI components, the development of AI components that support the AI model pipeline is more experimental with still limited methodological support and mostly done by data scientists and data engineers who are not familiar with software engineering. To bridge the methodological gap between AI and non-AI development, both AI team and non-AI team need to be clear about what exactly is being delivered by a project and share the same sprints and use a common co-versioning registry to track the progress~\cite{lu2022}. The close coupling of AI and non-AI development results in improved trust within the project team and better communication on both system-level and model-level ethical requirements. The challenge for the tight coupling might be that the non-AI component development is application centric while the AI component development is mostly data centric. There have been a few attempts in industry on continuously integrating AI components/models into the software, such as Microsoft Team Data Science Process~\footnote{\url{https://docs.microsoft.com/en-us/azure/architecture/data-science-process/overview}}, Amazon SageMaker Pipelines ~\footnote{\url{https://aws.amazon.com/sagemaker/pipelines/?nc1=h_ls}}, Azure Pipelines~\footnote{\url{https://www.azuredevopslabs.com/labs/vstsextend/aml/}}.

\subsubsection{Diverse team}
\hfill\\
AI pipelines are built by humans and thus may imply bias (such as racism and sexism) and produce discriminating results.  Also, the code of AI systems is written by developers who are primarily focused on technical aspects. 
Building a diverse project team can effectively eliminate bias and improve diversity and inclusion in AI systems~\cite{eitel2021beyond,luxton2014recommendations,zicari2021assessing}. The diversity can be across gender, race, age, sextual orientation, expertise, etc. 
A diverse team can drive creative thinking for greater innovation, but communication could become challenging due to different background and preference~\footnote{\url{https://futureofworking.com/11-advantages-and-disadvantages-of-diversity-in-the-workplace/}}. Google published 2022 Diversity Annual Report~\footnote{\url{https://about.google/belonging/at-work/}} which introduces the actions to build an inclusive workplace. Microsoft aims to integrate diversity and inclusion principles into their organization~\footnote{\url{https://careers.microsoft.com/us/en/diversityandinclusion}}. Meta has been working on creating diverse and inclusive work communities~\footnote{\url{https://www.workplace.com/diversity-and-inclusion}}.

\subsubsection{Stakeholder engagement}
\hfill\\
Stakeholders may have various ethical concerns about the development and use of AI systems. Keeping stakeholder engagement throughout the AI project is essential to building AI systems responsibly. Stakeholder engagement allows AI systems to better reflect their stakeholders’ needs and expectations~\cite{Shneiderman20,bilstrup2020staging,zicari2021co,zicari2021assessing}. There are various manners to engage stakeholders: interviews, online and offline meetings, project planning/review, participatory design workshops, crowd sourcing etc. Stakeholders may help the project team identify potential ethical risks before they become threats, but there maybe conflicting opinions from different stakeholders. Association for project management published ten stakeholder engagement principles~\footnote{\url{https://www.apm.org.uk/resources/find-a-resource/stakeholder-engagement/key-principles/}}. Australian Public Service Commission released stakeholder engagement guidelines~\footnote{\url{https://www.apsc.gov.au/initiatives-and-programs/workforce-information/taskforce-toolkit/stakeholder-engagement}}. Deloitte published a report on stakeholder engagement~\footnote{\url{https://www2.deloitte.com/content/dam/Deloitte/za/Documents/governance-risk-compliance/ZA_StakeholderEngagement_04042014.pdf}}.

\subsubsection{Continuous documentation using templates}
\hfill\\
Developers primarily focus on the code and often neglect updating the documentation during rapid iterations. The project teams need to create and continuously update documentations for the key artifacts of AI systems that may lead to ethical issues, such as data and models. Continuous documentation using templates helps track the evolution of artifacts and clarify the context in which AI systems are trustworthy~\cite{raji2020closing,jacovi2021formalizing,ahuja2020opening,hutchinson2021towards,zhou2022transparent,adkins2022prescriptive}. 
Google's model cards~\footnote{\url{https://modelcards.withgoogle.com/about}} enables transparent model reporting on model provenance and ethical evaluation~\cite{mitchell2019model,wadhwani2020machine}. Microsoft's datasheets for datasets~\footnote{\url{https://www.microsoft.com/en-us/research/project/datasheets-for-datasets/}} allows every dataset to be accompanied with a datasheet document~\cite{gebru2021datasheets}. IBM's AI service factsheets~\footnote{\url{https://www.ibm.com/blogs/research/2018/08/factsheets-ai/}} maintains AI services’ performance, safety, security, and provenance information~\cite{arnold2019factsheets}. Meta's method cards provides a prescriptive model specification templates that provides guidance on how to mitigate potential issues~\cite{adkins2022method,adkins2022prescriptive}.

\subsubsection{Failure mode and effects analysis (FMEA)}
\hfill\\
Ethical defects in AI systems are often detected through extensive simulation and testing in the later stages of development. However, this may lead to significant delays to timelines and additional development cost. FMEA is a bottom-up risk assessment method that can be used to identify ethical risks and calculate their priorities at the beginning of the development process~\cite{ebert2019validation}. FMEA was originally proposed in US Armed Forces Military Procedures document MIL-P-1629 in 1949 ~\footnote{\url{https://web.archive.org/web/20110722222459/https://assist.daps.dla.mil/quicksearch/basic_profile.cfm?ident_number=37027}}. Ford Motor Company firstly introduced FMEA to the automotive industry since mid 1970s ~\footnote{\url{https://fsp.portal.covisint.com/documents/106025/14555722/FMEA+Handbook+v4.2/4c14da5c-0842-4e60-a88b-75c18e143cf7?version=1.0}}. FMEA has been extended and adopted by Toyota’s Design Review Based on Failure Modes (DRBFM) ~\footnote{\url{https://www.sae.org/standards/content/j2886_201303/}} for assessing potential risk and reliability for Automotive and Non-Automotive applications. FMEA replies on experts to apply their professional knowledge and experience to the ethical risk assessment process.
Also, FMEA is better suited for bottom up analysis and not able to detect system-level complex ethical failures.

\subsubsection{Fault tree analysis (FTA)}
\hfill\\
Undesired system behaviors or decisions could lead to serious consequence and even cause loss of human lives. FTA~\cite{ebert2019validation} can be used to describe how system-level ethical failures are led by small ethical failure events through an analytical graph, i.e., fault tree. The development team can easily capture how ethical failures propagate in the AI system. Fault tree analysis can be done during the design or operation stage to anticipate the potential ethical risks and to recommend mitigation actions. FTA was firstly introduced by Bell Laboratories in 1962 to assess the safety of a missile launch control system\footnote{\url{https://www.osti.gov/servlets/purl/1315144}}. Boeing started using FTA to design civil aircrafts from 1966\footnote{\url{https://apps.dtic.mil/sti/citations/AD0847015}}. FTA was included in U.S. Army Materiel Command’s Engineering Design Handbook on Design for Reliability\footnote{\url{https://apps.dtic.mil/sti/pdfs/ADA026006.pdf}}. FTA assists in analyzing the ethical issues related to AI system artifacts and prioritizes the issues to address that contribute to an ethical risk. Hoever, it is complex to use for large system analysis, which may involve many ethical events and gates. Also, time can hardly be captured in FTA.

\subsubsection{Verifiable claim for AI system artifacts}
\hfill\\
The potential users of AI systems need methods for assessing an AI system’s ethical properties and compare the system to other systems. A verifiable claim platform can be built to support developers in making claims on ethical properties~\cite{gauerhof2020assuring} and conducting the verification~\cite{yap2020towards}. Such platform must consider the disparity of the stakeholder’s views. For example, developers might focus on reliability, while users might be interested in fairness. A verifiable claim is a statement about an AI system or an artifact (such as model or dataset) that is substantiated by a verification mechanism. The platform itself provides management capabilities such as claim creation and verification, access control, and dispute management. W3C Verifiable Claims Working Group aims to make expressing and exchanging claims~\footnote{\url{https://www.w3.org/2017/vc/WG/}}. The Open Web Application Security Project has published a Verifiable Claims documentation ~\footnote{\url{https://owasp.org/www-pdf-archive//OWASP-Austin-Mtg-2018Jan-CryptoParty-Dave-Sanford.pdf}}. The Ethereum Verifiable Claims is a method for off-chain variable claims\footnote{\url{https://eips.ethereum.org/EIPS/eip-1812\#ethereum-verifiable-claims}}.

\section{Process Patterns}
\begin{figure}
\centering
\includegraphics[width=\columnwidth]{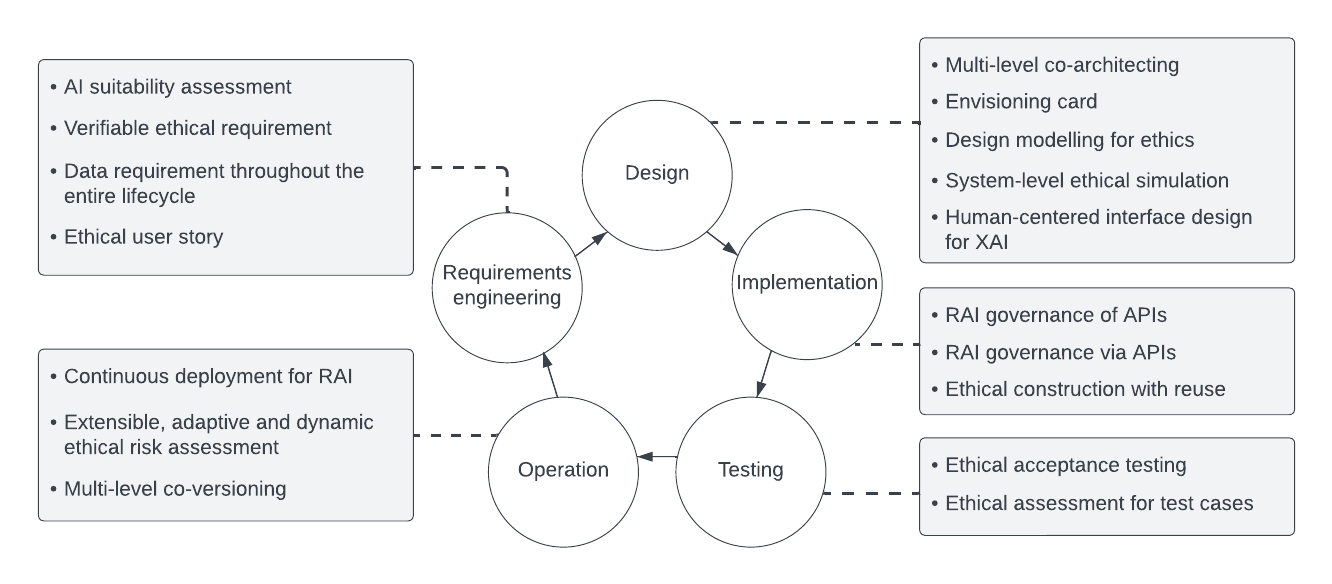}
\caption{Process patterns for responsible AI system development.} \label{fig:process}
\vspace{-2ex}
\end{figure}

In this section, we discuss the process patterns that can be incorporated into responsible AI system development processes. The process patterns are reusable methods and best practices which can be used by the development team during the development process.
Fig.~\ref{fig:process} illustrates the process patterns collected for each stage of the development process.

\subsection{Requirement engineering}

\subsubsection{AI suitability assessment}
\hfill\\
AI has a huge potential to provide effective solutions to tackle critical problems. However, it does not necessarily add value to every software system. Before starting to build a software system with AI, the development team first needs to identify the right problem to solve and the corresponding user needs. Once the problem is found and the environment where the system will be situated fully explored, the development team needs to analyse whether the system and the users benefit from AI or are they potentially degraded by AI~\cite{GooglePAIR_2021}. It is essential to make sure AI adds value to the design. Oftentimes, a heuristic-based design may be easier and cheaper to develop and may work better than an AI-based design in terms of predictability and transparency. AI suitability assessment can help the development team understand whether AI can add unique value to the design but may incur additional cost and require extra resources.

\subsubsection{Verifiable ethical requirement}
\hfill\\
The development of AI systems needs to adhere to AI ethics principles which are generally abstract and domain-agnostic. Ethical requirements need to be derived from the AI ethics principles to fit into a specific domain and system context~\cite{vogelsang2019requirements,zhu2022ai,horkoff2019non,bibal2021legal,perera2021impact}. Every ethical requirement specified in a requirements specification document should be put into a verifiable form (i.e., with acceptance criteria). This means a person or machine can later check that the AI system meets the ethical requirements that are derived from AI ethics principles and grounded in users' needs. Vague or unverifiable statements should be avoided~\cite{SWEBOK}. If there is no way to determine whether the AI system meets a particular ethical requirement, then this ethical requirement should be revised or removed. Ethical risk can be reduced via considering ethical requirements from the beginning of the development process and explicitly verifying ethical requirements. Some ethical principles/requirements may not be easily quantitatively validated~\cite{zhu2022ai}, such as human-centered values. There may be trade-offs between some ethical principles or requirements. The current practice to deal with the trade-offs is usually the developers following one principle while overwriting the others rather than building balanced trade-offs through patterns.

\subsubsection{Data requirements throughout the entire lifecycle}
\hfill\\
The quality of an AI model is largely dependent on the quality of the data used to train or evaluate. The lifecycle of data consists of several phases, including data collection, cleaning, preparation, validation, analysis, and termination. Unfortunately, the scope of data requirements~\cite{vogelsang2019requirements,Shneiderman20} often focuses on the data analysis phase and largely neglects the other key phases in the data lifecycle. This may lead to downstream ethical concerns such as AI model reliability, accountability, and fairness. AI systems can hardly be trusted when the data lifecycle is poorly managed. Data requirements need to be listed explicitly and specified throughout the data lifecycle (i.e., collection, cleaning, preparation, validation, analysis, and termination) taking into account ethical principles and involved stakeholders (i.e., data providers, data engineers, data scientists, data consumers, data auditors). Data requirements can be managed through data requirements specification. The specification could include detailed requirements for each phase in the data lifecycle, e.g., data collection requirements including data sources and collection methods. Google has created a template for dataset requirements specification~\cite{hutchinson2021towards}. 

\subsubsection{Ethical user story}
\hfill\\
Requirements elicitation methods are needed to collect detailed ethical requirements from stakeholders to capture AI ethics principles. In agile processes, ethical user stories~\cite{halme2021write,perera2021impact} can help the development team elicit ethical requirements for AI systems and implement AI ethics principles from the early stage of development. Ethical user stories are created to serve as items of the product backlog which is to be worked on by the development team in iterations (i.e., sprints). Card-based toolkits can be used to list questions related to AI ethics principles. The answers to those questions are integrated into ethical user stories to be included in sprint backlogs. The development team or users can write ethical user stories on cards or notes using predefined template and assign them to different sprints based on the priority. Ethical user stories make ethical requirements traceable both backward and forward, but they are difficult to scale for larger projects. Guide for Artificial Intelligence Ethical Requirements Elicitation\footnote{\url{https://josesiqueira.github.io/RE4AIEthicalGuide/index.html}} consists of 25 cards which are used by the development team to answer questions related to ethical principles. The answers are used to create ethical requirements in the form of ethical user stories which are included in sprint backlogs. ECCOLA~\cite{halme2021write} consists of 21 cards which are divided into 8 themes and with questions to be answered by the development team.

\subsection{Design}

\subsubsection{Multi-level co-architecting}
\hfill\\
Compared with traditional software, the architecture of AI systems is more complex due to different levels of integration. On the one hand, AI models are developed by data scientists/engineers via an AI model pipeline. The AI model pipeline usually comprises of a sequence of automatic steps including data collection, data cleaning, feature engineering, model training, and model evaluation. These steps can be viewed as software components for producing AI models from a software architecture perspective. On the other hand, the produced AI models cannot work alone and need to be integrated into software systems that are to be deployed in the real-world~\cite{muccini2021software}. The decisions made by the AI model need to be executed as actions via other software components. 
The architecture of an AI ecosystem consists of three layers: AI software supply chain, AI system, and operation infrastructure. The focus of AI software supply chain layer is about developing and managing AI and non-AI components~\cite{lo2021flra}, including AI model pipeline components, deployment components, co-versioning components, provenance tracking components, credential management components, etc. The AI system layer comprises AI components that embed AI models and non-AI components that use the outputs of AI components for overall system functionalities~\cite{lewis2021software}. The operation infrastructure layer is mainly about monitoring and feedback components. Multi-level co-architecting is required to ensure the seamless integration of different components, including co-architecting AI components and non-AI components and co-architecting of different AI model pipeline components. Multi-level co-architecting allows both system and model level requirements to be considered in design decision making.

\subsubsection{Envisioning card}
\hfill\\
AI ethics principles, including the human-centred values principles, are too high-level for developers who often lack the technical means to assure human values and ethics. Envisioning cards~\cite{umbrello2022role,bilstrup2020staging} are designed to help the development team operationalize human values during design processes of AI systems. The design of envisioning cards is based on four envisioning criteria, including stakeholder, time, value, and pervasiveness. The stakeholder criterion helps the development team takes into account the effects of an AI system on both direct stakeholders and indirect stakeholders. The time criterion emphasizes the long-term implication of AI systems on human, society, and environments. The value criterion guides the development team to consider the impact of AI systems on human values. The pervasiveness criterion discusses the challenges encountered if an AI system is widely adopted in terms of geography, culture, demographics, etc. The adoption of envisioning cards comes at a relatively low cost, in terms of both money and time. However, envisioning cards are hard to scale when the number of participants are large or the AI systems are complex.

\subsubsection{Design modelling for ethics}
\hfill\\
To reduce ethical risks, AI ethics principles need to be adhered to during the design process.
Design modelling methods can be extended and used to support the modelling of AI components and the ethical aspects, including: using UML to describe the architecture of AI systems and represent their ethical aspects~\cite{takeda2019accountable}, designing formal models taking into account human values~\cite{fish2021reflexive}, using ontologies to model the AI system artifacts for accountability~\cite{naja2021semantic,ayranci2022onml}, establishing RAI knowledge bases for making design decisions considering ethical concerns~\cite{sekiguchi2020organic}, using logic programming to implement ethical principles~\cite{anderson2018geneth}.
UML is an option to describe the AI systems and represent their ethical aspects~\cite{takeda2019accountable}. UML extension could be a declarative graphic notation for AI system architecture. Additional stereotypes/metamodel elements can be added for responsible-AI-by-design reference architecture (e.g., to describe AI pipeline components). Use case diagrams can help define the stakeholders and explain the functions they use, which are valuable for achieving accountability. State diagrams are useful to analyse the system states and identify the states that may cause ethical failures. Design patterns like AI mode switcher can take effect to change the state of an AI system to a more human controlled state. Sequence diagrams describe the human-AI interactions to ensure all the required explanations are provided. 
Using design modelling methods are helpful to capture and analyse ethical principles in design.
One disadvantage when using modelling languages is the time to create and manage the models. Also, the modelling languages do not scale up for large and complex systems.

\subsubsection{System-level ethical simulation}
\hfill\\
To avoid ethical disasters and gain public trust, it is necessary to model the real-world situations of AI systems without ethical risk. System-level simulation (e.g. ~\cite{singh2021simulation,ebert2019validation,dosovitskiy2017carla,regli2022anthropomorphic}), is a cost-effective way to imitate real-world situations and assess the behaviors of AI systems before deploying the AI systems in real-world. A simulation model needs to be built to mimic the possible behaviors and decisions of the AI system and assess the ethical impacts. The assessment results can be sent to the development team or potential users before the AI systems are deployed in the real-world. System-level simulation can predict potential ethical risks and avoid serious ethical disasters before deploying the AI systems in the real-world. However, the simulation model cannot represent all the behaviors and ethical impacts of AI systems in the real-world. The accuracy of assessment results is limited by the quality of the simulation model.

\subsubsection{Human-centered interface design for explainable AI}
\hfill\\
The end users often do not understand how decisions are made by AI systems and are not aware of the capabilities or limitations of the AI systems. The missing explainability may lead to a lack of trust and has been identified as one of the most urgent challenges of RAI to be addressed. Explainable AI (XAI) can be viewed as a human-AI interaction problem and achieved by effective human-centered interface design. 
Checklists (such as question bank) are often used to help design the explainable user interfaces~\cite{liao2020questioning,liao2021question, larasati2021ai} and understand the user needs, choices of XAI techniques, and XAI design factors~\cite{liao2021question}. For example, the checklist questions could consider the following aspects~\cite{liao2020questioning} for different stakeholders: input, output, how, performance (can be extended to ethical performance), why and why not, what if, etc. 
The design of conversational interfaces can be experimented via a Wizard of Oz study~\cite{jentzsch2019conversational}, in which users interact with a system that they believe to be autonomous but is actually being operated by a hidden human, called the Wizard. The conversation data is collected and analysed to understand requirements for a self-explanatory conversational interface.
There can be several ways to increase human trust in AI systems through human-centered user interface, including anthropomorphism~\cite{luxton2014recommendations}, proactive informing (such as capability/limitation of AI systems, ethics credentials, explanations of decisions/behaviors, potential outcomes, data use information).


\subsection{Implementation}
\subsubsection{RAI governance of APIs}
\hfill\\
APIs allow developers to solve problems more efficiently and can effectively reduce the development cost and time of AI systems. However, there may be ethical quality issues with APIs (e.g., data privacy breaches or fairness issues).
Ethical compliance checking for APIs is needed to detect if any ethics violation exists~\cite{hussain2020enterprise}. A knowledge-driven approach can be adopted to detect ethics issues through ethical knowledge graphs. Ethical knowledge graphs make meaningful entities and concepts, and their relationships in development of AI systems. With the ethical knowledge graph, the rich semantic relationships between entities are explicit and traceable across heterogeneous high-level documents and various AI systems artifacts. Ethical knowledge graphs can be built based on the ethical principles and guidelines (e.g., privacy knowledge graph based on GDPR~\cite{pandit2018towards,fan2020empirical}) and technical documents (e.g., API documentation) to support the ethical compliance checking for APIs. 

\subsubsection{RAI governance via APIs}
\hfill\\
Some AI systems may provide high risk capabilities, which can be used or modified to implement harmful tasks. To avoid harmful dual uses in AI systems~\cite{mitchell2022self}, developers should carefully design how their AI systems can be directly used and indirectly used (i.e., potential ways their systems can be adapted). Developers must restrict the way AI systems are used and preventing the users from getting around of restrictions by unauthorized reverse engineering or modification to the system design. Rather than fully opening the access to AI systems by allowing AI systems to run locally, developers could provide AI services on cloud and control the interactions with the AI services via APIs~\cite{shevlane2022structured}. For example, OpenAI's language model GPT-3 can be only integrated with AI systems through an API by approved users~\footnote{\url{https://openai.com/api/}}.
Google Vision AI limit its facial recognition feature to a few celebrities through API~\footnote{\url{https://cloud.google.com/vision}}.

\subsubsection{Ethical construction with reuse}
\hfill\\
Building AI systems from scratch can be very complex and time consuming. Very big companies usually have massive AI investments and large volumes of data to compete in the market, while smaller companies may only have a couple of data scientist and can hardly keep up with larger companies.
To speed up the development and reduce cost, it is highly desirable and valuable to reuse the AI artifacts (i.e., AI components and/or AI pipeline artifacts) across different applications. However, there might be ethical quality issues with the reused AI artifacts, which requires further assurance mechanisms. Ethical construction with reuse means to develop responsible AI systems with the use of existing AI artifacts that are compliant with AI ethics principles, e.g., from an organizational repository or an open-source platform. A marketplace can be built up to trade the reusable AI artifacts, including component code, models, and datasets. Blockchain can be adopted to design an immutable and transparent marketplace enabling the auction-based trading for AI artifacts and material assets (e.g., cloud resources)~\cite{six2021saiaas}. Ethics credentials might be required to be attached to the traded AI artifacts. Also, tooling support might be needed, such as model migration tool pytorch2keras\footnote{\url{https://github.com/gmalivenko/pytorch2keras}}) and glue code for compatibility
~\footnote{\url{https://insights.sei.cmu.edu/blog/software-engineering-for-machine-learning-characterizing-and-detecting-mismatch-in-machine-learning-systems/}}. Low/no code tools can also help to achieve ethical construction with reuse.

\subsection{Testing}

\subsubsection{Ethical acceptance testing}
\hfill\\
Since the AI Ethical principles are very high-level, they need to be captured through ethical requirements, which can be viewed as the agreed commitments by the development team and customers. 
Ethical acceptance testing (e.g., bias testing) is designed to detect the ethics-related design flaws and verify the ethical requirements (e.g. whether the data pipeline has appropriate privacy control, fairness testing for training/validation data)~\cite{chattopadhyay2021assessing,xie2020fairness,aggarwal2019black}. In agile process, the ethical requirements can be framed as ethical user stories and associated with ethical acceptance tests. The ethical acceptance tests are a contract between the customer and development team. The behavior of the AI system should be quantified by the acceptance tests and the acceptance criteria for each of the ethical principles should be defined in a testable way. The history of ethical acceptance testing should be recorded and tracked, such as how and by whom the ethical issues were fixed. A testing leader may be appointed to lead the ethical acceptance testing for each ethics principle. For example, when bias detected at runtime, the monitoring reports are returned to the bias testing leader~\cite{Shneiderman20,dixon2022principled}. Ethical acceptance tests capture the ethical requirements and measure how well the AI system meets ethical requirements, but may need to be amended frequently as ethical requirements change.

\subsubsection{Ethical assessment for test cases}
\hfill\\
The ethical quality assurance for AI systems is heavily dependent on ethical acceptance testing which is aimed at detecting and solving ethical issues in the AI system. A collection of test cases with expected results should be generated~\cite{murphy2007approach} and maintained for to detect possible ethical failures in a variety of extreme situations~\cite{goodall2014machine}. However, there might be ethical issues within the test cases. For example, the test data may introduce fairness or privacy issues~\cite{mehrabi2021survey}.
Preparing quality test cases is an integral part of ethical acceptance testing. A test case usually is composed of ID, description, preconditions, test steps, test data, expected results, actual results, status, creator name, creation date, executor name, execution date. All the test cases for verification and validation should pass the ethics assessment. This includes ethical risk assessment for test steps and test data. The creation and execution information are essential to track the accountability of ethical issues with test cases. Ethical assessment for test cases improves the ethical quality of the development process of AI systems, but new test cases need to be continually added and assessed when there is a new ethical requirement added or the operation context changes. 

\subsection{Operation}

\subsubsection{Continuous deployment for RAI}
\hfill\\
AI systems may frequently evolve due to their data dependency. When ethical performance degradation occurs over time, AI models need to be retrained with new data or features and reintegrated into AI components. The non-AI component may also need to be upgraded to meet new requirements or changing context. New versions of AI systems need to be frequently and continuously deployed into production environments. On the other hand, AI systems involve higher degree of uncertainty and risks associated with the autonomy of the AI systems. Thus, there is a strong desire for various deployment strategies to support continuous deployment~\cite{zhu2022ai,martinez2021developing}. 
There are various deployment strategies for AI systems. Phased deployment means deploying AI systems for a subset group of users initially to reduce ethical risk~\cite{hirsch2017designing}. The new version of AI systems rollouts incrementally and serves alongside the old version. Phased deployment can be also about automating decisions in phases to better supervise and control automation. This usually depends on the stakes of the situations and the level of confidence that users may have with automatic decisions made by AI systems. Further, A/B testing deployment\cite{john2020architecting} is a common deployment strategy undertaken in industry, where different versions of the AI model deployed to production. The models are compared and selected based on their ethical performance. In addition, the existing reliability practices, like redundancy, are also applicable to AI components in an AI system. Multiple AI models work independently to improve the ethical performance of the AI components. 
Applying various deployment strategies helps to reduce the ethical risk.
Users can be quickly redirected to the older version or the other version of AI systems/models.
However, it is complex and expensive to adopt different deployment strategies during operations.

\subsubsection{Extensible, adaptive and dynamic ethical risk assessment}
\hfill\\
The current risk-based approach to ethical principles is often a done-once-and-forget type of algorithm-level risk assessment~\cite{schiff2020principles,Shneiderman20,martin2017taking,zicari2021z,henderson2021certifai} and mitigation for a subset of ethical principles (such as privacy or fairness~\footnote{\url{https://www.nist.gov/artificial-intelligence/proposal-identifying-and-managing-bias-artificial-intelligence-sp-1270}}) at a particular development step (such as Canada's Algorithmic Impact Assessment Tool\footnote{\url{https://www.canada.ca/en/government/system/digital-government/digital-government-innovations/responsible-use-ai/algorithmic-impact-assessment.html}}), which is not sufficient for the highly uncertain and continual learning AI systems. In addition, the context of AI systems varies with the application domains, organizations, culture, and regions.
It is essential to perform continuous risk assessment and mitigation of responsible AI systems~\cite{staples2016continuous,d2018towards}. The ethical risk assessment framework can be built with guided extension points for different context (e.g. culture context). The risk mitigation can be designed from three aspects: reducing frequency occurrence, consequence size, and consequence response. 
Extensible, adaptive and dynamic risk assessment can effectively ensure an AI system adheres to AI ethics principles throughout the whole lifecycle, but it might be hard to measure some of the ethical principles, e.g., human-centered values.

\subsubsection{Multi-level co-versioning}
\hfill\\
AI systems involve two levels of relationships and dependencies across various AI artifacts, including supply chain level and system level. 
At the system level, there are multiple versions of AI components and non-AI components. At the supply chain level, there are different versions of data, model, code, and configuration, which are used to produce different versions of AI components~\cite{lewis2021software}. 
At the system level, the AI components that embed AI models are integrated into AI systems and interact with non-AI components. On the other hand, the retraining of AI models introduces new versions of data, code and configuration parameters. If federated learning is adopted, for each round of training, a global model is ensembled based on local models sent from participating clients~\cite{lo2022architectural}. It is important to capture all these dependencies during the development process. 
Multi-level co-versioning provides end-to-end traceability and accountability throughout the whole lifecycle of AI systems, but the collection and documentation of co-versioning information incur additional development cost. There have been many version control tools in industry focusing on supply chain level co-versioning, e.g., MLflow Model Registry on Databricks~\footnote{\url{https://docs.databricks.com/applications/mlflow/model-registry.html}} and
Amazon provenance tool\footnote{\url{https://www.amazon.science/publications/automatically-tracking-metadata-and-provenance-of-machine-learning-experiments}}, and Data Version Control (DVC)\footnote{\url{https://dvc.org/}}.

\section{Product Patterns}
\begin{figure}
\centering
\includegraphics[width=\columnwidth]{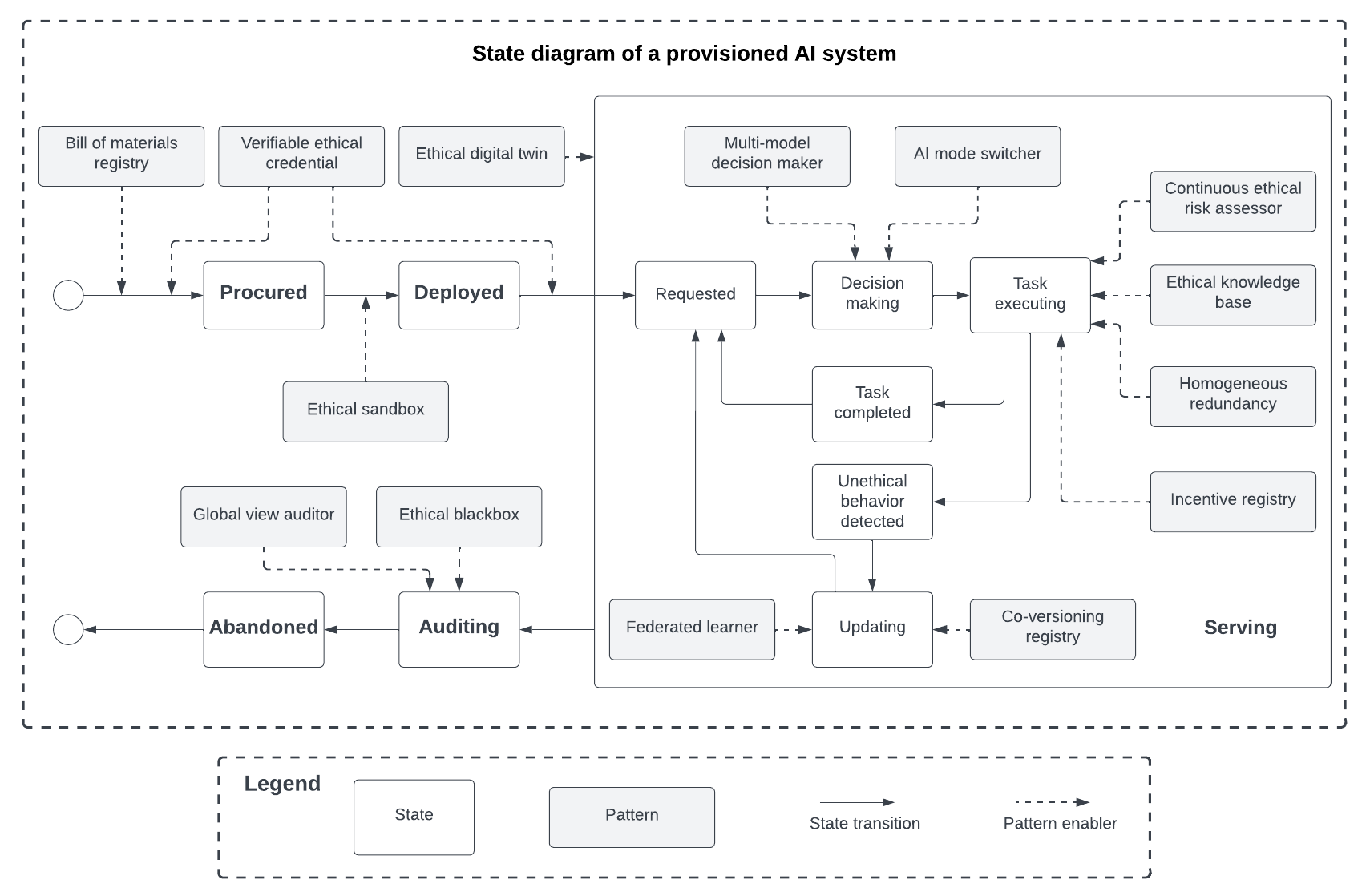}
\caption{Product patterns for responsible-AI-by-design.} \label{fig:product}
\vspace{-2ex}
\end{figure}

This section provides a system-level guidance on how to design the architecture of responsible AI systems. We present a collection of product patterns (i.e., design patterns) for building responsible-AI-by-design into AI systems.
Broadly, an AI system is comprised by three layers: (1) the supply chain layer that generates the software components which compose the AI system, (2) the system layer which is deployed AI system, and (3) the operation infrastructure layer that provides auxiliary functions to the AI system. Fig.~\ref{fig:product} presents the identified products patterns for each of the three layers. Those product patterns can be embedded into the AI ecosystems as product features. Fig.~\ref{fig:product} illustrates a state diagram of a provisioned AI system and highlights the patterns associating with relevant states or transitions, which show when the product patterns could take effect. 

\begin{figure}
\centering
\includegraphics[width=0.85\columnwidth]{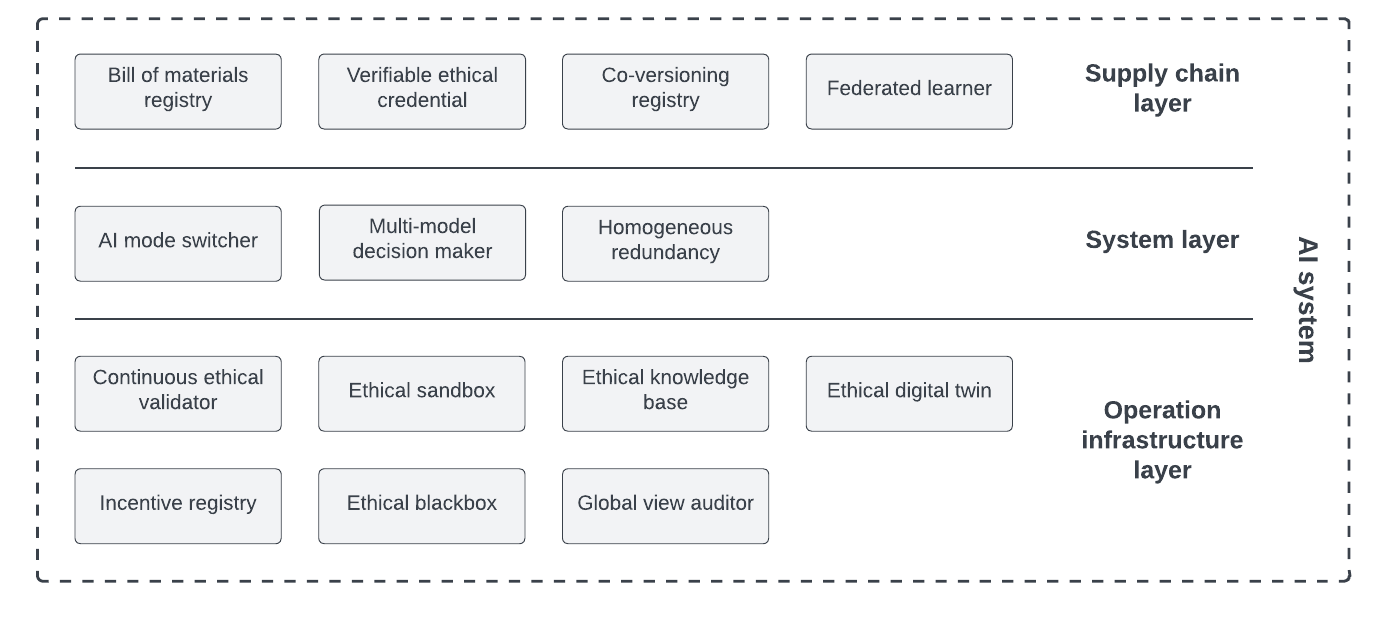}
\caption{Product patterns for responsible-AI-by-design architecture of an AI system.} \label{fig:eco}
\vspace{-2ex}
\end{figure}

\subsection{Supply chain patterns}
\subsubsection{Bill of materials registry}
\hfill\\
Bill of materials registry~\cite{sbom_minimum_elements_report_2021, barclay2019towards} can be designed to keep a formal machine-readable record of the supply chain details of the components used in building an AI system, including component name, version, supplier, dependency relationship, author, and timestamp. In addition to supply chain details of the components, context documents (like model cards~\cite{mitchell2019model} for reporting AI models, and datasheets for the datasets~\cite{gebru2021datasheets} used to train AI models) can also be integrated to the bill of materials registry. 
The main purpose of bill of materials registry is to provide traceability and transparency into the components within AI systems so that ethical issues can be tracked and addressed~\cite{barclay2022providing}. Some platforms manage bill of materials registry, such as OpenBOM~\footnote{\url{https://www.openbom.com/}}, Codenotary~\footnote{\url{https://codenotary.com/}}, Snorkel Flow~\footnote{\url{https://snorkel.ai/}}. Immutable data infrastructure can store the bill of materials to enable integrity. For example, the manufacturers of autonomous vehicles could maintain a material registry contract on blockchain to track their components' supply chain information, e.g., the version and supplier of the third-party navigation component.
Stakeholders can access the supply chain details of each component of interest in AI systems via bill of materials registry. As AI systems evolve over time, the bill of materials may need to be updated frequently. The cost of managing the bill of materials of all the components depends on the complexity of the AI system.

\subsubsection{Verifiable ethical credential}
\hfill\\
Verifiable ethical credentials can be used as evidence of ethical compliance for AI systems, components, models, developers, operators\footnote{\url{https://certnexus.com/certification/ceet/}}, users, organizations, and development processes~\cite{chu2022decentralized,luxton2014recommendations,paulk1993capability}. Verifiable credentials are data that could be cryptographically verified and be presented with strong proofs~\cite{world2019verifiable}. Publicly accessible data infrastructure needs to be built to support the generation and verification of the ethical credentials on a neutral platform.
Before using AI systems, users may verify the systems' ethical credential to check if the systems are compliant with AI ethics principles or regulations~\cite{chu2022decentralized}. On the other hand, the users may be required to provide the ethical credentials to use and operate the AI systems, e.g., to ensure the flight safety of drones. 
Verifiable ethical credential helps increase user trust towards an AI system through conferring the trust that the user has with the authority that issues the credential to AI systems, organizations that develop AI systems and the operators that operate AI systems. Such transitive trust relationship is critical in the efficient functioning of the AI system.
With an ethical credential, an AI system could provide proof of compliance as an incentive for the users to use the AI system, thus increase AI adoption.
Ethical credential may be forged, which makes the verification of authenticity of the ethical credentials becomes challenging. Blockchain could be adopted to build the credential infrastructure to ensure data integrity. For example,	Securekey~\footnote{\url{https://securekey.com/}} is a blockchain-based infrastructure for ID management with support of verifiable credential.

\subsubsection{Co-versioning registry}
\hfill\\
Compared with traditional software, AI systems involve different levels of dependencies and may evolve more frequently due to their data-dependent behaviors. From the viewpoint of the AI system, it is important to know the version of the AI component integrated into the system. From the viewpoint of the AI component, it is important to know what datasets and parameters were used to train the AI model and what data was used to evaluate the AI model.
Co-versioning of the components or AI artifacts of AI systems provides end-to-end provenance guarantees across the entire lifecycle of AI systems. Co-versioning registry can track the co-evolution of components or AI artifacts~\cite{lewis2021software,lo2022architectural}. There are different levels of co-versioning: co-versioning of AI components and non-AI components, co-versioning of the artifacts within the AI components (i.e., co-versioning of data, model, code, configurations, and co-versioning of local models and global models in federated learning). Co-versioning enables effective maintenance and evolution of AI component because the deployed model or code can be traced to the exact set of artifacts, parameters and metadata that were used to develop the model and code.  MLflow Model Registry~\footnote{\url{https://docs.databricks.com/applications/mlflow/model-registry.html}} is a model repository and set of APIs that enable management of the full lifecycle of MLflow Models, including model lineage and versioning.  

\subsubsection{Federated learner}
\hfill\\
Despite the widely deployed mobile or IoT devices generating massive amounts of data, lack of training data is still a challenge for AI systems given the increasing concern in data privacy.
Federated learner trains an AI model across multiple edge devices or servers with local data samples. Federated learner~\cite{lo2021flra,lo2022architectural,bonawitz2017practical,suzen2020novel,bennati2017primal,verachtert2021privacy,sugianto2021privacy,lo2021blockchain} preserves the data privacy by training models locally on the client devices and formulating a global model on a central server based on the local model updates, e.g., train the visual perception model locally in each vehicle. Decentralized learning is a variant of federated learning, which could use blockchain to remove the single point of failure and coordinate the learning process in a fully decentralized way~\cite{warnat2021swarm}. TensorFlow Federated~\footnote{\url{https://www.tensorflow.org/federated}} is an open-source framework for machine learning on decentralized data sources. FATE~\footnote{\url{https://fate.fedai.org/}} is an open-source project that support the federated AI ecosystem.

\subsection{System patterns}

\subsubsection{AI mode switcher}
\hfill\\
When to use AI at decision making points can be a major architectural design decision when designing an AI system. Adding an AI mode switcher to the AI system offers users efficient invocation and dismissal mechanisms for activating and deactivating the AI component whenever needed, thus, defer the architectural decision to the execution time which is decided by the end user or the operator of the AI system. AI mode switcher is like a kill switch of AI system that could immediately shut down the AI component and thus, stop its negative effects~\cite{vassilakopoulou2020sociotechnical,hax_2022,GooglePAIR_2021}, e.g., turning off the automated driving system and disconnecting it from the internet. The decisions made by the AI component can be executed automatically or reviewed by a human expert before being executed in critical situations. The human expert serves to approve or override the decisions (e.g., skipping the path generated by the navigation system). Human intervention can also happen after acting the AI decision through the fallback mechanism that reverses the system back to the state before executing the AI decision. A built-in guard can be used to ensure that the AI component is only activated within the predefined conditions (such as domain of use, boundaries of competence). The end users or the operators can ask questions or report complaints/failures/near misses through a recourse channel after observing a bad decision from AI component. Tesla autopilot~\footnote{\url{https://www.tesla.com/autopilot}} has multiple driver assistance features that can be enabled or disabled during the driving. Users maintain control of the vehicles and can override the operations by these features at runtime.
Baidu autonomous mini-bus Robobus \footnote{\url{https://apollo.auto/minibus/}} requires a staff in the seat to supervise the self-driving operations, and the bus can be switched to manual driving mode by braking.

\subsubsection{Multi-model decision maker}
\hfill\\
In reliability community of software system, traditional architecture-based software reliability is based on software component. The existing reliability practices, like redundancy, are also applicable to AI components in an AI system. In addition, reasonable combination of multiple AI models that are normally work independently could improve the performance (e.g., accuracy) of the AI component. 
Multi-model decision-maker employs different models to perform the same task or enable a single decision, e.g., deploying different algorithms for visual perception. It improves the reliability by deploying different models under different context (e.g., different geo-location regions) and enabling fault-tolerance by cross-validating ethical requirements for a single decision~\cite{dai2021more,nafreen2020architecture}. Different consensus protocols could be defined to make the final decision, for example, taking the majority decision. Another strategy is to only accept the same results from the employed models. In addition, the end user or the operator could step in to review the output from the multiple models and make a final decision based on human’s expertise. Scikit-learn~\footnote{\url{https://github.com/scikit-learn/scikit-learn}} is a Python package that supports using multiple learning al-gorithms to obtain better performance through ensemble learning. 
AWS Fraud Detection Using Machine Learning solution trains an unsupervised anomaly detection model in addition to a supervised model, to augment the prediction results~\footnote{\url{https://aws.amazon.com/solutions/implementations/fraud-detection-using-machine-learning/}}.
IBM Watson Natural Language Understanding  uses an ensemble learning framework to include predictions from multiple emotion detection models~\footnote{\url{https://www.ibm.com/au-en/cloud/watson-natural-language-understanding}}.

\subsubsection{Homogeneous redundancy}
\hfill\\
N-version programming is a software design pattern to ensure fault tolerance of software~\cite{knight2002n}. Similarly, deploying multiple redundant and identical AI components (e.g., two brake control components) can be a solution to tolerate the individual AI component with high uncertainty that may make unethical decisions or the individual adversary hardware component that produces malicious data or behaves unethically~\cite{nafreen2020architecture}. A cross-check can be conducted for the outputs provided by multiple components of a single type. The results are accepted only there is a consensus among the redundant components. The results that are not accepted automatically according to a consensus protocol can be further reviewed by the end user or the operator of the AI system. Waymo~\footnote{\url{https://waymo.com/}} contains multiple redundant components at various levels, including redundant braking, steering, and inertial measurement systems for vehicle positioning.

\subsection{Operation infrastructure patterns}

\subsubsection{Continuous ethical validator}
\hfill\\
AI components of an AI system often require continual learning based on new data collected during operation of the AI system. Continuous ethical validator deployed in an AI system continuously monitors and validates the outcomes of AI components (e.g., the path recommended by the navigation system) against the ethical requirements~\cite{staples2016continuous,john2020architecting}.
The outcomes of AI systems are about whether the AI system provides the intended benefits and behaves appropriately given the situation. The time and frequency of validation can be configured. Version-based feedback and rebuild alert are sent when the predefined conditions regarding the ethical requirement are met. 
AWS SageMaker Model Monitor~\footnote{\url{https://docs.aws.amazon.com/sagemaker/latest/dg/model-monitor.html}} continuously monitors the bias drift of the AI models in production. 
Qualdo~\footnote{\url{https://www.qualdo.ai/monitor-ml-model-performance-monitoring/}}  is an AI monitoring solution that monitors data quality and model drift. 
Azure Machine Learning~\footnote{\url{https://docs.microsoft.com/en-us/azure/machine-learning/monitor-azure-machine-learning}} uses Azure Monitor to create monitoring data. Azure Monitor is a full stack monitoring service.

\subsubsection{Ethical sandbox}
\hfill\\
Given AI systems are of high stake, it is risky to run the entire system in the same execution environment. Ethical sandbox can be applied to isolate an AI component from other AI components and non-AI components by running the AI component separately in a safe environment~\cite{lavaei2021towards}, e.g. sandboxing the unverified visual perception component. Thus, the AI component could execute without affecting other components and the output of the AI system. 
Ethical sandbox is an emulated environment with no access to the rest of the AI system. An emulation environment duplicates all the hardware and software functionality of an AI system. Thus, developers could run an AI component safely to determine how it works and whether it is responsible before widely deploying the AI component. Maximal tolerable probability of violating the ethical requirements should be defined as ethical margin for the sandbox. 
A watch dog can be used to limit the execution time of the AI component to reduce the ethical risk, e.g., only activating the visual perception component for 5 mins on the bridges built especially for autonomous vehicles. Fastcase AI Sandbox~\footnote{\url{https://www.fastcase.com/sandbox/}} provides a secure platform for the users to upload dataset and do data analysis in a safe environment. AI Sandbox~\footnote{\url{https://aisandbox.dev/}} provides an AI execution and RESTful interface that could be used by modern programming languages. 

\subsubsection{Ethical knowledge base}
\hfill\\

The ecosystem of AI systems involves broad ethical knowledge, such as AI ethics principles, regulations, and guidelines. Such ethical knowledge is scattered and is usually implicit or abstract to end users or even developers and data scientists who primarily without legal background and focus more on the technical aspects of AI systems. Ethical knowledge base, such as a knowledge graph, makes meaningful entities and concepts, and their relationships in design, implementation, deployment, and operation of AI systems~\cite{naja2021semantic,sekiguchi2020organic,esnaolaontology}. With the ethical knowledge based, the rich semantic relationships between entities are explicit and traceable across heterogeneous high-level documents on one hand, and different artifacts across the AI system lifecycle on the other hand. Thus, ethical requirements of the AI system can be systematically accessed and analyzed using the ethical knowledge base. 
Awesome AI guidelines~\footnote{\url{https://github.com/EthicalML/awesome-artificial-intelligence-guidelines}} aims to provide a mapping between ecosystem of guidelines, principles, codes of ethics, standards and regulation around artificial intelligence.
The responsible AI community portal~\footnote{\url{https://portal.responsible.ai/}}  is provided by AI Global, which is an evolving repository of reports, standards, models, government policies, datasets, and open-source software to inform and support responsible AI development. 
Responsible AI Knowledge-base~\footnote{\url{https://github.com/alexandrainst/responsible-ai}} is a knowledge base of different areas of using and developing AI in a responsible way.

\subsubsection{Ethical digital twin}
\hfill\\
Simulation is designed to imitate a real-world situation. Before running AI system in real-world, it is important to perform system-level simulation through an ethical digital twin running on a simulation infrastructure to understand the behaviors of the AI system and assess ethical risks in a cost-effective way. Digital twin~\cite{shafto2012modeling} is introduced by NASA as a digital representation of a real system used in lab-testing activities. The digital twin of an AI system could be used to represent the behaviors of the AI system and forecast change impacts. Ethical digital twin can also be used during operation of the AI system to assess the system's runtime behaviors and decisions based on the simulation model using the real-time data. The assessment results can be sent back to alert the system or user before the unethical behavior or decision takes effect~\cite{dosovitskiy2017carla}. Vehicle manufacturers can use the ethical digital twin to explore the limits of autonomous vehicles based on the collected real-time data, such as NVIDIA DRIVE Sim~\footnote{\url{https://developer.nvidia.com/drive/drive-sim}} and	rfPro~\footnote{\url{https://www.rfpro.com/}}.

\subsubsection{Incentive registry}
\hfill\\
Incentive mechanisms are effective treatments in motivating AI systems and encouraging the stakeholders involved in the AI system ecosystem to execute tasks in a responsible manner. An incentive registry records the rewards that correspond to the AI system's ethical behavior and outcome of decisions~\cite{weng2019deepchain,zhang2020blockchain}, e.g., rewards for path planning without ethical risks. There are various ways to formulate the incentive mechanism, for example, using reinforcement learning, or building the incentive mechanism on a publicly accessible data infrastructure like blockchain~\cite{zhang2020blockchain}. 
Traditional incentive mechanisms for human participants include reputation-based 
and payment-based. However, it is challenging to formulate the form of rewards in the context of responsible AI as the ethical impact of AI systems' decisions and behaviors might hardly to be measured for some of the ethical principles (such as human values). Furthermore, the incentive mechanism needs to be agreed by all the stakeholders who may have different views on the ethical impact. In addition, there may be trade-offs between different principles, which makes the design harder. The Open Science Rewards and Incentives Registry~\footnote{\url{https://openscienceregistry.org/}} incentivizes the development of an academic career structure that fosters outputs, practices and behaviors to maximize contributions to a shared research knowledge system. 
FLoBC~\footnote{\url{https://github.com/Oschart/FLoBC}} is a tool for federated learning over blockchain which utilizes a reward/punishment policy to incentivize legitimate training, and to punish and hinder malicious trainers.

\begin{figure}
\centering
\includegraphics[width=\columnwidth]{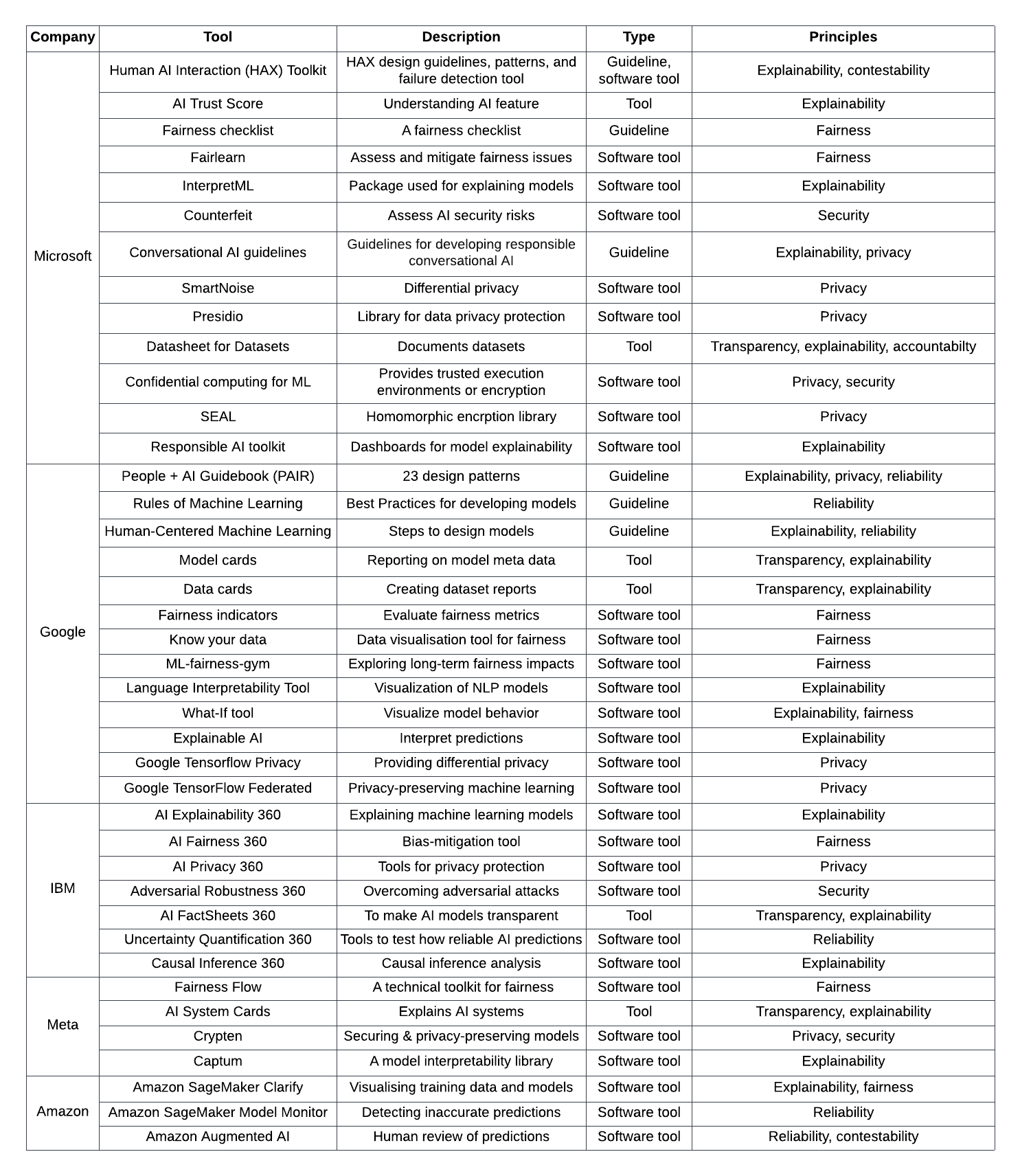}
\caption{Top 5 major industry players on responsible AI according to the number of tools.} \label{fig:products}
\vspace{-2ex}
\end{figure}

\subsubsection{Ethical blackbox}
\hfill\\
Black box was introduced initially for aircraft several decades ago for recording critical flight data. The intention of adding a black box to aircrafts is to collect evidence of the actions of system and the surrounding context information for analysis after near misses and failures. The near misses and failures are specific to the use cases. Although the primary usage of a black box is accident investigation, black boxes are useful for other purposes. Data collection and the analysis could support improvement of the system. The purpose of embedding an ethical black box in an AI system is to investigate why and how an AI system caused an accident or a near miss. The ethical black box continuously records sensor data, internal status data, decisions, behaviors (both system and operator) and effects~\cite{falco2020distributedblack,winfield2017case,falco2021governing}. For example, an ethical black box could be built into the automated driving system to record the behaviors of the system and driver and their effects~\cite{falco2020distributedblack}. All these data need to be kept as evidence with the timestamp and location data. Designing the ethical black box is challenging as the ethical metrics need to be identified for data collection. Also, design decisions need to be made on what data should be recorded and where the data should be stored (e.g. using a blockchain-based immutable log or a cloud-based data storage). RoBoTIPS~\footnote{\url{https://www.robotips.co.uk/}} aims to develop an ethical black box for social robots, to enable the explainability of their behavior.

\subsubsection{Global view auditor}
\hfill\\
When an accident happens, there might be more than one AI systems or multiple AI components within on AI system involved (e.g., multiple autonomous vehicles in an accident). The data collected from each involved AI systems/components might conflict with each other since the individual AI system/component may have their own perception.
Global-view auditor is a component that collects information from multiple AI components/AI systems, process the information to identify discrepancies among the information collected~\cite{miguel2021putting}. Based on the result, the global-view auditor may alert the AI system/component with wrong perception, thus, avoid negative impacts or identify liability when negative events occur.
This pattern can be also used to improve the decision-making of an AI system by taking the knowledge from other systems. For example, an autonomous vehicle may increase their visibility using the perceptions of others to make better decisions at runtime.
Global-view auditor enables accountability that cover different perceptions of AI components/systems that are involved and redresses the conflicting information collected from multiple AI components/ systems. 

\section{Related Work}

The challenge of responsible AI has attracted significant attention in both industry and academia.
To achieve responsible AI, there have been nearly 100 high-level AI ethics principles and guidelines issued by governments, organizations, and companies~\cite{jobin2019global}.
Some degree of consensus around AI ethics principles has been achieved~\cite{fjeld2020principled}. A principle-based approach allows technology-independent operationalization of responsible AI. However, these principles are very abstract and high-level for stakeholders of AI systems to use in practice. 

Significant efforts have been put on algorithm-level solutions which mainly focus on a subset of principles. Fig.~\ref{fig:products} lists the top 5 major industry players on responsible AI according to their number of responsible AI tools based on the results of our MLR study. Most of these tools focus on privacy, security, reliability, safety, fairness, and explainability from an AI model perspective. More work is needed on transparency, accountability, contestability, human-centered values, and human, societal, and environmental wellbeing, particularly from a system perspective.

Overall, AI ethics principles need to be operationalized in the form of concrete patterns and best practices that are usable by AI developers and other stakeholders to build up responsible AI systems. 
Some add-hoc sets of guidebooks, question banks, checklists and templates have started to appear.
Microsoft's Human AI Interaction (HAX) Toolkit provides a set of HAX guidelines and patterns~\footnote{\url{https://www.microsoft.com/en-us/haxtoolkit/ai-guidelines/}}. However, those guidelines and patterns only focus on interaction design and do not provide any guidance on development and governance. Google's People AI Guidebook (PAIR) summaries 23 design patterns~\footnote{\url{https://pair.withgoogle.com/guidebook/patterns}} which mainly address some of the AI ethics principles for AI models, including explainability, privacy, reliability. Process and governance guidelines are not discussed in Google's PAIR. Although OECD provides a framework of tools for trustworthy AI~\cite{oecd},
the framework largely contains categorised but disjointed software tools,
lacking process-related linkages.
Thus, a systematic and operationalized guidance for AI system stakeholders is required throughout the entire lifecycle of AI systems.

There have been a few survey papers on operationalizing responsible AI~\cite{schiff2020principles,smit2020review,anagnostou2022characteristics}. However, the findings and insights in these papers are still around principles and do not provide concrete and actionable guidelines for stakeholders to use in practice. Our previous roadmap paper~\cite{lu2022} discusses the current state and identify the critical research challenges in the area of software engineering for responsible AI based on an initial systematic literature review. This pattern catalogue paper is built on top of the published roadmap and provides a comprehensive list of concrete patterns from multi-level governance patterns to process and product patterns based on the results of a multivocal literature review. In the Responsible AI Pattern Catalogue, we present structured knowledge about the patterns, including context, problem, solution, benefits, drawbacks, and known uses. The full version of the Responsible AI Pattern Catalogue is available online\textsuperscript{\ref{catalogue}}.

\section{Threats to Validity}
External Validity - First, "responsible AI" is loosely defined with many terms that refer to emerging technologies in this area. There is a set of terms currently being used in the community to mean largely the same thing: responsible AI, AI ethics, ethical AI, trustworthy AI, and trust in AI. This issue has been addressed by including search terms that are being used interchangeably in the search string to ensure that all the relevant work were covered. Another issue is that many solutions were initially designed only for addressing one of the AI ethics principles but could be identified as a pattern and extended to implement responsible AI. To mitigate this threat, we included all the AI ethics principles in the search string as supplementary terms.

Internal Validity - To mitigate the threat of not finding all relevant studies, we performed a rigorous search using defined keywords and executed snowballing that allows us to recover the missing studies from the literature. To address the bias, one researcher performed the screening of titles, abstracts and full-texts. The other researcher evaluated a random sample of the selected studies after screening to check the consistency of their inclusion/exclusion decisions.

\section{Conclusion}
To operationalize responsible AI, this paper adopts a pattern-oriented approach and presents a comprehensive responsible AI pattern catalogue that AI system stakeholders can utilise to ensure the developed AI systems are trustworthy throughout the entire governance and engineering lifecycle, from multi-level governance patterns to concrete process and product patterns. These patterns offer a systematic and actionable system-level guidance with consequence analysis and well-known uses for AI system stakeholders to reference during the governance and development processes. We are currently building a Question Bank and a software tool for AI risk assessment, which will use the responsible AI pattern catalogue as one of the knowledge sources to recommend mitigation strategies. We also plan to validate the utility, usability and effectiveness of the pattern catalogue and the supporting tools in industrial projects.

\input{main.bbl}



\end{document}

%% file: main.bbl

%% file: main.bbl
\begin{thebibliography}{100}
\providecommand{\url}[1]{#1}
\csname url@samestyle\endcsname
\providecommand{\newblock}{\relax}
\providecommand{\bibinfo}[2]{#2}
\providecommand{\BIBentrySTDinterwordspacing}{\spaceskip=0pt\relax}
\providecommand{\BIBentryALTinterwordstretchfactor}{4}
\providecommand{\BIBentryALTinterwordspacing}{\spaceskip=\fontdimen2\font plus
\BIBentryALTinterwordstretchfactor\fontdimen3\font minus
  \fontdimen4\font\relax}
\providecommand{\BIBforeignlanguage}[2]{{%
\expandafter\ifx\csname l@#1\endcsname\relax
\typeout{** WARNING: IEEEtran.bst: No hyphenation pattern has been}%
\typeout{** loaded for the language `#1'. Using the pattern for}%
\typeout{** the default language instead.}%
\else
\language=\csname l@#1\endcsname
\fi
#2}}
\providecommand{\BIBdecl}{\relax}
\BIBdecl

\bibitem{jobin2019global}
A.~Jobin, M.~Ienca, and E.~Vayena, ``The global landscape of ai ethics
  guidelines,'' \emph{Nature Machine Intelligence}, vol.~1, no.~9, pp.
  389--399, 2019.

\bibitem{fjeld2020principled}
J.~Fjeld~et. al., ``Principled artificial intelligence: Mapping consensus in
  ethical and rights-based approaches to principles for ai,'' \emph{Berkman
  Klein Center Research Publication}, no. 2020-1, 2020.

\bibitem{liao2020questioning}
\BIBentryALTinterwordspacing
Q.~V. Liao, D.~Gruen, and S.~Miller, ``Questioning the ai: Informing design
  practices for explainable ai user experiences,'' in \emph{Proceedings of the
  2020 CHI Conference on Human Factors in Computing Systems}, ser. CHI
  '20.\hskip 1em plus 0.5em minus 0.4em\relax New York, NY, USA: Association
  for Computing Machinery, 2020, p. 1–15. [Online]. Available:
  \url{https://doi.org/10.1145/3313831.3376590}
\BIBentrySTDinterwordspacing

\bibitem{liao2021question}
Q.~V. Liao, M.~Pribi{\'c}, J.~Han, S.~Miller, and D.~Sow, ``Question-driven
  design process for explainable ai user experiences,'' \emph{arXiv preprint
  arXiv:2104.03483}, 2021.

\bibitem{larasati2021ai}
R.~Larasati, A.~De~Liddo, and E.~Motta, ``Ai healthcare system interface:
  Explanation design for non-expert user trust,'' in \emph{ACMIUI-WS 2021:
  Joint Proceedings of the ACM IUI 2021 Workshops}, vol. 2903.\hskip 1em plus
  0.5em minus 0.4em\relax CEUR Workshop Proceedings, 2021.

\bibitem{han2022checklist}
S.-H. Han and H.-J. Choi, ``Checklist for validating trustworthy ai,'' in
  \emph{2022 IEEE International Conference on Big Data and Smart Computing
  (BigComp)}.\hskip 1em plus 0.5em minus 0.4em\relax IEEE, 2022, pp. 391--394.

\bibitem{raji2020closing}
I.~D. Raji, A.~Smart, R.~N. White, M.~Mitchell, T.~Gebru, B.~Hutchinson,
  J.~Smith-Loud, D.~Theron, and P.~Barnes, ``Closing the ai accountability gap:
  Defining an end-to-end framework for internal algorithmic auditing,'' in
  \emph{Proceedings of the 2020 conference on fairness, accountability, and
  transparency}, 2020, pp. 33--44.

\bibitem{jacovi2021formalizing}
A.~Jacovi, A.~Marasovi{\'c}, T.~Miller, and Y.~Goldberg, ``Formalizing trust in
  artificial intelligence: Prerequisites, causes and goals of human trust in
  ai,'' in \emph{Proceedings of the 2021 ACM conference on fairness,
  accountability, and transparency}, 2021, pp. 624--635.

\bibitem{ahuja2020opening}
M.~K. Ahuja, M.-B. Belaid, P.~Bernab{\'e}, M.~Collet, A.~Gotlieb, C.~Lal,
  D.~Marijan, S.~Sen, A.~Sharif, and H.~Spieker, ``Opening the software
  engineering toolbox for the assessment of trustworthy ai,'' \emph{arXiv
  preprint arXiv:2007.07768}, 2020.

\bibitem{hutchinson2021towards}
B.~Hutchinson, A.~Smart, A.~Hanna, E.~Denton, C.~Greer, O.~Kjartansson,
  P.~Barnes, and M.~Mitchell, ``Towards accountability for machine learning
  datasets: Practices from software engineering and infrastructure,'' in
  \emph{Proceedings of the 2021 ACM Conference on Fairness, Accountability, and
  Transparency}, 2021, pp. 560--575.

\bibitem{zhou2022transparent}
Z.~Zhou, Z.~Li, Y.~Zhang, and L.~Sun, ``Transparent-ai blueprint: Developing a
  conceptual tool to support the design of transparent ai agents,''
  \emph{International Journal of Human--Computer Interaction}, pp. 1--28, 2022.

\bibitem{adkins2022prescriptive}
D.~Adkins, B.~Alsallakh, A.~Cheema, N.~Kokhlikyan, E.~McReynolds, P.~Mishra,
  C.~Procope, J.~Sawruk, E.~Wang, and P.~Zvyagina, ``Prescriptive and
  descriptive approaches to machine-learning transparency,'' in \emph{CHI
  Conference on Human Factors in Computing Systems Extended Abstracts}, 2022,
  pp. 1--9.

\bibitem{Beck1987}
K.~Beck and W.~Cunningham, ``Using pattern languages for object oriented
  programs,'' in \emph{Conference on Object-Oriented Programming, Systems,
  Languages, and Applications (OOPSLA)}, 1987.

\bibitem{kitchenham2007guidelines}
B.~A. Kitchenham and S.~Charters, ``Guidelines for performing systematic
  literature reviews in software engineering,'' Tech. Rep., 2007.

\bibitem{GAROUSI2019101}
\BIBentryALTinterwordspacing
V.~Garousi, M.~Felderer, and M.~V. Mäntylä, ``Guidelines for including grey
  literature and conducting multivocal literature reviews in software
  engineering,'' \emph{Information and Software Technology}, vol. 106, pp.
  101--121, 2019. [Online]. Available:
  \url{https://www.sciencedirect.com/science/article/pii/S0950584918301939}
\BIBentrySTDinterwordspacing

\bibitem{DISER_2020}
{DISER (Australian Government)}, ``{Australia's AI Ethics Principles},''
  \url{https://industry.gov.au/data-and-publications/australias-artificial-intelligence-ethics-framework/australias-ai-ethics-principles},
  2020, accessed: 17 Aug 2022.

\bibitem{martin1997pattern}
R.~C. Martin, D.~Riehle, and F.~Buschmann, \emph{Pattern languages of program
  design 3}.\hskip 1em plus 0.5em minus 0.4em\relax Addison-Wesley Longman
  Publishing Co., Inc., 1997.

\bibitem{Shneiderman20}
B.~Shneiderman, ``Bridging the gap between ethics and practice: Guidelines for
  reliable, safe, and trustworthy human-centered ai systems,'' \emph{ACM Trans.
  Interact. Intell. Syst.}, vol.~10, no.~4, 2020.

\bibitem{shneiderman2021responsible}
------, ``Responsible ai: Bridging from ethics to practice,''
  \emph{Communications of the ACM}, vol.~64, no.~8, pp. 32--35, 2021.

\bibitem{jackson2021ethics}
B.~R. Jackson, Y.~Ye, J.~M. Crawford, M.~J. Becich, S.~Roy, J.~R. Botkin, M.~E.
  de~Baca, and L.~Pantanowitz, ``The ethics of artificial intelligence in
  pathology and laboratory medicine: principles and practice,'' \emph{Academic
  Pathology}, vol.~8, p. 2374289521990784, 2021.

\bibitem{schaich2021four}
J.~Schaich~Borg, ``Four investment areas for ethical ai: Transdisciplinary
  opportunities to close the publication-to-practice gap,'' \emph{Big Data \&
  Society}, vol.~8, no.~2, p. 20539517211040197, 2021.

\bibitem{papagiannidis2021deploying}
E.~Papagiannidis, I.~M. Enholm, C.~Dremel, P.~Mikalef, and J.~Krogstie,
  ``Deploying ai governance practices: A revelatory case study,'' in
  \emph{Conference on e-Business, e-Services and e-Society}.\hskip 1em plus
  0.5em minus 0.4em\relax Springer, 2021, pp. 208--219.

\bibitem{ibanez2021operationalising}
J.~C. Ib{\'a}{\~n}ez and M.~V. Olmeda, ``Operationalising ai ethics: how are
  companies bridging the gap between practice and principles? an exploratory
  study,'' \emph{AI \& SOCIETY}, pp. 1--25, 2021.

\bibitem{dignum2019ensuring}
V.~Dignum, ``Ensuring responsible ai in practice,'' in \emph{Responsible
  Artificial Intelligence}.\hskip 1em plus 0.5em minus 0.4em\relax Springer,
  2019, pp. 93--105.

\bibitem{zhobe2021magic}
A.~Zhobe, H.~Jahankhani, R.~Fong, P.~Elevique, and H.~Baajour, ``The magic
  quadrant: Assessing ethical maturity for artificial intelligence,'' in
  \emph{Cybersecurity, Privacy and Freedom Protection in the Connected
  World}.\hskip 1em plus 0.5em minus 0.4em\relax Springer, 2021, pp. 313--326.

\bibitem{fukas2021developing}
P.~Fukas, J.~Rebstadt, F.~Remark, and O.~Thomas, ``Developing an artificial
  intelligence maturity model for auditing.'' in \emph{ECIS}, 2021.

\bibitem{alsheiabni2019towards}
S.~Alsheiabni, Y.~Cheung, and C.~Messom, ``Towards an artificial intelligence
  maturity model: from science fiction to business facts,'' 2019.

\bibitem{henriksen2021situated}
A.~Henriksen, S.~Enni, and A.~Bechmann, ``Situated accountability: Ethical
  principles, certification standards, and explanation methods in applied ai,''
  in \emph{Proceedings of the 2021 AAAI/ACM Conference on AI, Ethics, and
  Society}, 2021, pp. 574--585.

\bibitem{martin2017taking}
C.~D. Martin and T.~T. Makoundou, ``Taking the high road ethics by design in
  ai,'' \emph{ACM Inroads}, vol.~8, no.~4, pp. 35--37, 2017.

\bibitem{yap2020towards}
R.~H. Yap, ``Towards certifying trustworthy machine learning systems,'' in
  \emph{International Workshop on the Foundations of Trustworthy AI Integrating
  Learning, Optimization and Reasoning}.\hskip 1em plus 0.5em minus 0.4em\relax
  Springer, 2020, pp. 77--82.

\bibitem{cihon2021ai}
P.~Cihon, M.~J. Kleinaltenkamp, J.~Schuett, and S.~D. Baum, ``Ai certification:
  Advancing ethical practice by reducing information asymmetries,'' \emph{IEEE
  Transactions on Technology and Society}, vol.~2, no.~4, pp. 200--209, 2021.

\bibitem{boza2021implementing}
P.~Boza and T.~Evgeniou, ``Implementing ai principles: Frameworks, processes,
  and tools,'' 2021.

\bibitem{luxton2014recommendations}
D.~D. Luxton, ``Recommendations for the ethical use and design of artificial
  intelligent care providers,'' \emph{Artificial intelligence in medicine},
  vol.~62, no.~1, pp. 1--10, 2014.

\bibitem{mccradden2020patient}
M.~D. McCradden, S.~Joshi, J.~A. Anderson, M.~Mazwi, A.~Goldenberg, and
  R.~Zlotnik~Shaul, ``Patient safety and quality improvement: Ethical
  principles for a regulatory approach to bias in healthcare machine
  learning,'' \emph{Journal of the American Medical Informatics Association},
  vol.~27, no.~12, pp. 2024--2027, 2020.

\bibitem{zhu2022ai}
L.~Zhu, X.~Xu, Q.~Lu, G.~Governatori, and J.~Whittle, ``Ai and
  ethics—operationalizing responsible ai,'' in \emph{Humanity Driven
  AI}.\hskip 1em plus 0.5em minus 0.4em\relax Springer, 2022, pp. 15--33.

\bibitem{sloane2022german}
M.~Sloane and J.~Zakrzewski, ``German ai start-ups and “ai ethics”: Using a
  social practice lens for assessing and implementing socio-technical
  innovation,'' in \emph{2022 ACM Conference on Fairness, Accountability, and
  Transparency}, 2022, pp. 935--947.

\bibitem{hooker2018toward}
J.~N. Hooker and T.~W.~N. Kim, ``Toward non-intuition-based machine and
  artificial intelligence ethics: A deontological approach based on modal
  logic,'' in \emph{Proceedings of the 2018 AAAI/ACM Conference on AI, Ethics,
  and Society}, 2018, pp. 130--136.

\bibitem{d2018towards}
M.~d'Aquin, P.~Troullinou, N.~E. O'Connor, A.~Cullen, G.~Faller, and L.~Holden,
  ``Towards an" ethics by design" methodology for ai research projects,'' in
  \emph{Proceedings of the 2018 AAAI/ACM Conference on AI, Ethics, and
  Society}, 2018, pp. 54--59.

\bibitem{benjamins2019responsible}
R.~Benjamins, A.~Barbado, and D.~Sierra, ``Responsible ai by design in
  practice,'' \emph{arXiv preprint arXiv:1909.12838}, 2019.

\bibitem{eitel2021beyond}
R.~Eitel-Porter, ``Beyond the promise: implementing ethical ai,'' \emph{AI and
  Ethics}, vol.~1, no.~1, pp. 73--80, 2021.

\bibitem{lee2021risk}
M.~S.~A. Lee and J.~Singh, ``Risk identification questionnaire for detecting
  unintended bias in the machine learning development lifecycle,'' in
  \emph{Proceedings of the 2021 AAAI/ACM Conference on AI, Ethics, and
  Society}, 2021, pp. 704--714.

\bibitem{redmill2002risk}
F.~Redmill, ``Risk analysis\-a subjective process,'' \emph{Engineering
  Management Journal}, vol.~12, no.~2, pp. 91--96, 2002.

\bibitem{schultz2022towards}
M.~D. Schultz and P.~Seele, ``Towards ai ethics’ institutionalization:
  knowledge bridges from business ethics to advance organizational ai ethics,''
  \emph{AI and Ethics}, pp. 1--13, 2022.

\bibitem{barclay2019towards}
I.~Barclay, A.~Preece, I.~Taylor, and D.~Verma, ``Towards traceability in data
  ecosystems using a bill of materials model,'' in \emph{International Workshop
  on Science Gateways}.\hskip 1em plus 0.5em minus 0.4em\relax CEUR-WS, 2019.

\bibitem{sbom_mini}
NTIA, ``The {Minimum} {Elements} {For} a {Software} {Bill} of {Materials}
  {(SBOM)},''
  \url{https://www.ntia.doc.gov/files/ntia/publications/sbom_minimum_elements_report.pdf},
  2021, accessed 18 Aug 2022.

\bibitem{kasenberg2018norms}
D.~Kasenberg, T.~Arnold, and M.~Scheutz, ``Norms, rewards, and the intentional
  stance: Comparing machine learning approaches to ethical training,'' in
  \emph{Proceedings of the 2018 AAAI/ACM Conference on AI, Ethics, and
  Society}, 2018, pp. 184--190.

\bibitem{sand2022responsibility}
M.~Sand, J.~M. Dur{\'a}n, and K.~R. Jongsma, ``Responsibility beyond design:
  Physicians’ requirements for ethical medical ai,'' \emph{Bioethics},
  vol.~36, no.~2, pp. 162--169, 2022.

\bibitem{amershi2019software}
S.~Amershi, A.~Begel, C.~Bird, R.~DeLine, H.~Gall, E.~Kamar, N.~Nagappan,
  B.~Nushi, and T.~Zimmermann, ``Software engineering for machine learning: A
  case study,'' in \emph{2019 IEEE/ACM 41st International Conference on
  Software Engineering: Software Engineering in Practice (ICSE-SEIP)}.\hskip
  1em plus 0.5em minus 0.4em\relax IEEE, 2019, pp. 291--300.

\bibitem{hussain2022can}
W.~Hussain, M.~Shahin, R.~Hoda, J.~Whittle, H.~Perera, A.~Nurwidyantoro, R.~A.
  Shams, and G.~Oliver, ``How can human values be addressed in agilemethods a
  case study on safe,'' \emph{IEEE Transactions on Software Engineering}, 2022.

\bibitem{lu2022}
Q.~Lu, L.~Zhu, X.~Xu, J.~Whittle, and Z.~Xing, ``Towards a roadmap on software
  engineering for responsible ai,'' in \emph{2022 IEEE/ACM 1st International
  Conference on AI Engineering – Software Engineering for AI (CAIN)}, 2022,
  pp. 101--112.

\bibitem{zicari2021assessing}
R.~V. Zicari, J.~Brusseau, S.~N. Blomberg, H.~C. Christensen, M.~Coffee, M.~B.
  Ganapini, S.~Gerke, T.~K. Gilbert, E.~Hickman, E.~Hildt \emph{et~al.}, ``On
  assessing trustworthy ai in healthcare. machine learning as a supportive tool
  to recognize cardiac arrest in emergency calls,'' \emph{Frontiers in Human
  Dynamics}, p.~30, 2021.

\bibitem{bilstrup2020staging}
K.-E.~K. Bilstrup, M.~H. Kaspersen, and M.~G. Petersen, ``Staging reflections
  on ethical dilemmas in machine learning: A card-based design workshop for
  high school students,'' in \emph{Proceedings of the 2020 ACM Designing
  Interactive Systems Conference}, 2020, pp. 1211--1222.

\bibitem{zicari2021co}
R.~V. Zicari, S.~Ahmed, J.~Amann, S.~A. Braun, J.~Brodersen, F.~Bruneault,
  J.~Brusseau, E.~Campano, M.~Coffee, A.~Dengel \emph{et~al.}, ``Co-design of a
  trustworthy ai system in healthcare: deep learning based skin lesion
  classifier,'' \emph{Frontiers in Human Dynamics}, vol.~3, p. 688152, 2021.

\bibitem{mitchell2019model}
M.~Mitchell, S.~Wu, A.~Zaldivar, P.~Barnes, L.~Vasserman, B.~Hutchinson,
  E.~Spitzer, I.~D. Raji, and T.~Gebru, ``Model cards for model reporting,'' in
  \emph{Proceedings of the conference on fairness, accountability, and
  transparency}, 2019, pp. 220--229.

\bibitem{wadhwani2020machine}
A.~Wadhwani and P.~Jain, ``Machine learning model cards transparency review:
  Using model card toolkit,'' in \emph{2020 IEEE Pune Section International
  Conference (PuneCon)}.\hskip 1em plus 0.5em minus 0.4em\relax IEEE, 2020, pp.
  133--137.

\bibitem{gebru2021datasheets}
T.~Gebru, J.~Morgenstern, B.~Vecchione, J.~W. Vaughan, H.~Wallach, H.~D. Iii,
  and K.~Crawford, ``Datasheets for datasets,'' \emph{Communications of the
  ACM}, vol.~64, no.~12, pp. 86--92, 2021.

\bibitem{arnold2019factsheets}
M.~Arnold, R.~K. Bellamy, M.~Hind, S.~Houde, S.~Mehta, A.~Mojsilovi{\'c},
  R.~Nair, K.~N. Ramamurthy, A.~Olteanu, D.~Piorkowski \emph{et~al.},
  ``Factsheets: Increasing trust in ai services through supplier's declarations
  of conformity,'' \emph{IBM Journal of Research and Development}, vol.~63, no.
  4/5, pp. 6--1, 2019.

\bibitem{adkins2022method}
D.~Adkins, B.~Alsallakh, A.~Cheema, N.~Kokhlikyan, E.~McReynolds, P.~Mishra,
  C.~Procope, J.~Sawruk, E.~Wang, and P.~Zvyagina, ``Method cards for
  prescriptive machine-learning transparency,'' in \emph{2022 IEEE/ACM 1st
  International Conference on AI Engineering--Software Engineering for AI
  (CAIN)}.\hskip 1em plus 0.5em minus 0.4em\relax IEEE, 2022, pp. 90--100.

\bibitem{ebert2019validation}
C.~Ebert and M.~Weyrich, ``Validation of autonomous systems,'' \emph{IEEE
  Software}, vol.~36, no.~5, pp. 15--23, 2019.

\bibitem{gauerhof2020assuring}
L.~Gauerhof, R.~Hawkins, C.~Picardi, C.~Paterson, Y.~Hagiwara, and I.~Habli,
  ``Assuring the safety of machine learning for pedestrian detection at
  crossings,'' in \emph{International Conference on Computer Safety,
  Reliability, and Security}.\hskip 1em plus 0.5em minus 0.4em\relax Springer,
  2020, pp. 197--212.

\bibitem{GooglePAIR_2021}
G.~PAIR, ``People + ai guidebook,'' \url{pair.withgoogle.com/guidebook}, 2021,
  accessed: 17 Aug 2022.

\bibitem{vogelsang2019requirements}
A.~Vogelsang and M.~Borg, ``Requirements engineering for machine learning:
  Perspectives from data scientists,'' in \emph{2019 IEEE 27th International
  Requirements Engineering Conference Workshops (REW)}.\hskip 1em plus 0.5em
  minus 0.4em\relax IEEE, 2019, pp. 245--251.

\bibitem{horkoff2019non}
J.~Horkoff, ``Non-functional requirements for machine learning: Challenges and
  new directions,'' in \emph{2019 IEEE 27th international requirements
  engineering conference (RE)}.\hskip 1em plus 0.5em minus 0.4em\relax IEEE,
  2019, pp. 386--391.

\bibitem{bibal2021legal}
A.~Bibal, M.~Lognoul, A.~De~Streel, and B.~Fr{\'e}nay, ``Legal requirements on
  explainability in machine learning,'' \emph{Artificial Intelligence and Law},
  vol.~29, no.~2, pp. 149--169, 2021.

\bibitem{perera2021impact}
H.~Perera, R.~Hoda, R.~A. Shams, A.~Nurwidyantoro, M.~Shahin, W.~Hussain, and
  J.~Whittle, ``The impact of considering human values during requirements
  engineering activities,'' \emph{arXiv preprint arXiv:2111.15293}, 2021.

\bibitem{SWEBOK}
I.~Society, P.~Bourque, and R.~Fairley, ``Guide to the software engineering
  body of knowledge (swebok (r)),'' 2014.

\bibitem{halme2021write}
E.~Halme, V.~Vakkuri, J.~Kultanen, M.~Jantunen, K.-K. Kemell, R.~Rousi, and
  P.~Abrahamsson, ``How to write ethical user stories? impacts of the eccola
  method,'' in \emph{International Conference on Agile Software
  Development}.\hskip 1em plus 0.5em minus 0.4em\relax Springer, Cham, 2021,
  pp. 36--52.

\bibitem{muccini2021software}
H.~Muccini and K.~Vaidhyanathan, ``Software architecture for ml-based systems:
  what exists and what lies ahead,'' in \emph{2021 IEEE/ACM 1st Workshop on AI
  Engineering-Software Engineering for AI (WAIN)}.\hskip 1em plus 0.5em minus
  0.4em\relax IEEE, 2021, pp. 121--128.

\bibitem{lo2021flra}
S.~K. Lo, Q.~Lu, H.-Y. Paik, and L.~Zhu, ``Flra: A reference architecture for
  federated learning systems,'' in \emph{European Conference on Software
  Architecture}.\hskip 1em plus 0.5em minus 0.4em\relax Springer, 2021, pp.
  83--98.

\bibitem{lewis2021software}
G.~A. Lewis, I.~Ozkaya, and X.~Xu, ``Software architecture challenges for ml
  systems,'' in \emph{2021 IEEE International Conference on Software
  Maintenance and Evolution (ICSME)}.\hskip 1em plus 0.5em minus 0.4em\relax
  IEEE, 2021, pp. 634--638.

\bibitem{umbrello2022role}
S.~Umbrello, ``The role of engineers in harmonising human values for ai systems
  design,'' \emph{Journal of Responsible Technology}, vol.~10, p. 100031, 2022.

\bibitem{takeda2019accountable}
M.~Takeda, Y.~Hirata, Y.-H. Weng, T.~Katayama, Y.~Mizuta, and A.~Koujina,
  ``Accountable system design architecture for embodied ai: a focus on physical
  human support robots,'' \emph{Advanced Robotics}, vol.~33, no.~23, pp.
  1248--1263, 2019.

\bibitem{fish2021reflexive}
B.~Fish and L.~Stark, ``Reflexive design for fairness and other human values in
  formal models,'' in \emph{Proceedings of the 2021 AAAI/ACM Conference on AI,
  Ethics, and Society}, 2021, pp. 89--99.

\bibitem{naja2021semantic}
I.~Naja, M.~Markovic, P.~Edwards, and C.~Cottrill, ``A semantic framework to
  support ai system accountability and audit,'' in \emph{European Semantic Web
  Conference}.\hskip 1em plus 0.5em minus 0.4em\relax Springer, 2021, pp.
  160--176.

\bibitem{ayranci2022onml}
P.~Ayranci, P.~Lai, N.~Phan, H.~Hu, A.~Kolinowski, D.~Newman, and D.~Dou,
  ``Onml: an ontology-based approach for interpretable machine learning,''
  \emph{Journal of Combinatorial Optimization}, pp. 1--24, 2022.

\bibitem{sekiguchi2020organic}
K.~Sekiguchi and K.~Hori, ``Organic and dynamic tool for use with knowledge
  base of ai ethics for promoting engineers’ practice of ethical ai design,''
  \emph{AI \& SOCIETY}, vol.~35, no.~1, pp. 51--71, 2020.

\bibitem{anderson2018geneth}
M.~Anderson and S.~L. Anderson, ``Geneth: A general ethical dilemma analyzer,''
  \emph{Paladyn, Journal of Behavioral Robotics}, vol.~9, no.~1, pp. 337--357,
  2018.

\bibitem{singh2021simulation}
V.~Singh, S.~K.~S. Hari, T.~Tsai, and M.~Pitale, ``Simulation driven design and
  test for safety of ai based autonomous vehicles,'' in \emph{Proceedings of
  the IEEE/CVF Conference on Computer Vision and Pattern Recognition}, 2021,
  pp. 122--128.

\bibitem{dosovitskiy2017carla}
A.~Dosovitskiy, G.~Ros, F.~Codevilla, A.~Lopez, and V.~Koltun, ``Carla: An open
  urban driving simulator,'' in \emph{Conference on robot learning}.\hskip 1em
  plus 0.5em minus 0.4em\relax PMLR, 2017, pp. 1--16.

\bibitem{regli2022anthropomorphic}
C.~Regli and B.~Annighoefer, ``An anthropomorphic approach to establish an
  additional layer of trustworthiness of an ai pilot,'' in \emph{Software
  Engineering 2022 Workshops}.\hskip 1em plus 0.5em minus 0.4em\relax
  Gesellschaft f{\"u}r Informatik eV, 2022.

\bibitem{jentzsch2019conversational}
S.~F. Jentzsch, S.~H{\"o}hn, and N.~Hochgeschwender, ``Conversational
  interfaces for explainable ai: a human-centred approach,'' in
  \emph{International Workshop on Explainable, Transparent Autonomous Agents
  and Multi-Agent Systems}.\hskip 1em plus 0.5em minus 0.4em\relax Springer,
  2019, pp. 77--92.

\bibitem{hussain2020enterprise}
F.~Hussain, R.~Hussain, B.~Noye, and S.~Sharieh, ``Enterprise api security and
  gdpr compliance: Design and implementation perspective,'' \emph{IT
  Professional}, vol.~22, no.~5, pp. 81--89, 2020.

\bibitem{pandit2018towards}
H.~J. Pandit, D.~O'Sullivan, and D.~Lewis, ``Towards knowledge-based systems
  for gdpr compliance.'' in \emph{CKGSemStats@ ISWC}, 2018.

\bibitem{fan2020empirical}
M.~Fan, L.~Yu, S.~Chen, H.~Zhou, X.~Luo, S.~Li, Y.~Liu, J.~Liu, and T.~Liu,
  ``An empirical evaluation of gdpr compliance violations in android mhealth
  apps,'' in \emph{2020 IEEE 31st international symposium on software
  reliability engineering (ISSRE)}.\hskip 1em plus 0.5em minus 0.4em\relax
  IEEE, 2020, pp. 253--264.

\bibitem{mitchell2022self}
E.~Mitchell, P.~Henderson, C.~D. Manning, D.~Jurafsky, and C.~Finn,
  ``Self-destructing models: Increasing the costs of harmful dual uses in
  foundation models,'' in \emph{First Workshop on Pre-training: Perspectives,
  Pitfalls, and Paths Forward at ICML 2022}, 2002.

\bibitem{shevlane2022structured}
T.~Shevlane, ``Structured access to ai capabilities: an emerging paradigm for
  safe ai deployment,'' \emph{arXiv preprint arXiv:2201.05159}, 2022.

\bibitem{six2021saiaas}
N.~Six, A.~Perrichon-Chr{\'e}tien, and N.~Herbaut, ``Saiaas: A blockchain-based
  solution for secure artificial intelligence as-a-service,'' in \emph{The
  International Conference on Deep Learning, Big Data and Blockchain}.\hskip
  1em plus 0.5em minus 0.4em\relax Springer, 2021, pp. 67--74.

\bibitem{chattopadhyay2021assessing}
A.~Chattopadhyay, A.~Ali, and D.~Thaxton, ``Assessing the alignment of social
  robots with trustworthy ai design guidelines: A preliminary research study,''
  in \emph{Proceedings of the Eleventh ACM Conference on Data and Application
  Security and Privacy}, 2021, pp. 325--327.

\bibitem{xie2020fairness}
W.~Xie and P.~Wu, ``Fairness testing of machine learning models using deep
  reinforcement learning,'' in \emph{2020 IEEE 19th International Conference on
  Trust, Security and Privacy in Computing and Communications
  (TrustCom)}.\hskip 1em plus 0.5em minus 0.4em\relax IEEE, 2020, pp. 121--128.

\bibitem{aggarwal2019black}
A.~Aggarwal, P.~Lohia, S.~Nagar, K.~Dey, and D.~Saha, ``Black box fairness
  testing of machine learning models,'' in \emph{Proceedings of the 2019 27th
  ACM Joint Meeting on European Software Engineering Conference and Symposium
  on the Foundations of Software Engineering}, 2019, pp. 625--635.

\bibitem{dixon2022principled}
R.~B.~L. Dixon, ``A principled governance for emerging ai regimes: lessons from
  china, the european union, and the united states,'' \emph{AI and Ethics}, pp.
  1--18, 2022.

\bibitem{murphy2007approach}
C.~Murphy, G.~E. Kaiser, and M.~Arias, ``An approach to software testing of
  machine learning applications,'' 2007.

\bibitem{goodall2014machine}
N.~J. Goodall, ``Machine ethics and automated vehicles,'' in \emph{Road vehicle
  automation}.\hskip 1em plus 0.5em minus 0.4em\relax Springer, 2014, pp.
  93--102.

\bibitem{mehrabi2021survey}
N.~Mehrabi, F.~Morstatter, N.~Saxena, K.~Lerman, and A.~Galstyan, ``A survey on
  bias and fairness in machine learning,'' \emph{ACM Computing Surveys (CSUR)},
  vol.~54, no.~6, pp. 1--35, 2021.

\bibitem{martinez2021developing}
S.~Mart{\'\i}nez-Fern{\'a}ndez, X.~Franch, A.~Jedlitschka, M.~Oriol, and
  A.~Trendowicz, ``Developing and operating artificial intelligence models in
  trustworthy autonomous systems,'' in \emph{International Conference on
  Research Challenges in Information Science}.\hskip 1em plus 0.5em minus
  0.4em\relax Springer, 2021, pp. 221--229.

\bibitem{hirsch2017designing}
T.~Hirsch, K.~Merced, S.~Narayanan, Z.~E. Imel, and D.~C. Atkins, ``Designing
  contestability: Interaction design, machine learning, and mental health,'' in
  \emph{Proceedings of the 2017 Conference on Designing Interactive Systems},
  2017, pp. 95--99.

\bibitem{john2020architecting}
M.~M. John, H.~Holmstr{\"o}m~Olsson, and J.~Bosch, ``Architecting ai
  deployment: A systematic review of state-of-the-art and state-of-practice
  literature,'' in \emph{International Conference on Software Business}.\hskip
  1em plus 0.5em minus 0.4em\relax Springer, 2020, pp. 14--29.

\bibitem{schiff2020principles}
D.~Schiff, B.~Rakova, A.~Ayesh, A.~Fanti, and M.~Lennon, ``Principles to
  practices for responsible ai: closing the gap,'' \emph{arXiv preprint
  arXiv:2006.04707}, 2020.

\bibitem{zicari2021z}
R.~V. Zicari, J.~Brodersen, J.~Brusseau, B.~D{\"u}dder, T.~Eichhorn, T.~Ivanov,
  G.~Kararigas, P.~Kringen, M.~McCullough, F.~M{\"o}slein \emph{et~al.},
  ``Z-inspection{\textregistered}: a process to assess trustworthy ai,''
  \emph{IEEE Transactions on Technology and Society}, vol.~2, no.~2, pp.
  83--97, 2021.

\bibitem{henderson2021certifai}
J.~Henderson, S.~Sharma, A.~Gee, V.~Alexiev, S.~Draper, C.~Marin, Y.~Hinojosa,
  C.~Draper, M.~Perng, L.~Aguirre \emph{et~al.}, ``Certifai: a toolkit for
  building trust in ai systems,'' in \emph{Proceedings of the Twenty-Ninth
  International Conference on International Joint Conferences on Artificial
  Intelligence}, 2021, pp. 5249--5251.

\bibitem{staples2016continuous}
M.~Staples, L.~Zhu, and J.~Grundy, ``Continuous validation for data analytics
  systems,'' in \emph{Proceedings of the 38th International Conference on
  Software Engineering Companion}, 2016, pp. 769--772.

\bibitem{lo2022architectural}
S.~K. Lo, Q.~Lu, L.~Zhu, H.-y. Paik, X.~Xu, and C.~Wang, ``Architectural
  patterns for the design of federated learning systems,'' \emph{Journal of
  Systems and Software}, vol. 191, p. 111357, 2022.

\bibitem{sbom_minimum_elements_report_2021}
T.~U. S.~D. of~Commerce, ``The minimum elements for a software bill of
  materials (sbom),''
  \url{https://www.ntia.doc.gov/files/ntia/publications/sbom_minimum_elements_report.pdf},
  2021, accessed 17 Aug 2022.

\bibitem{barclay2022providing}
I.~Barclay, A.~Preece, I.~Taylor, S.~K. Radha, and J.~Nabrzyski, ``Providing
  assurance and scrutability on shared data and machine learning models with
  verifiable credentials,'' \emph{Concurrency and Computation: Practice and
  Experience}, p. e6997, 2022.

\bibitem{chu2022decentralized}
W.~Chu, ``A decentralized approach towards responsible ai in social
  ecosystems,'' in \emph{Proceedings of the International AAAI Conference on
  Web and Social Media}, vol.~16, 2022, pp. 79--89.

\bibitem{paulk1993capability}
M.~C. Paulk, B.~Curtis, M.~B. Chrissis, and C.~V. Weber, ``Capability maturity
  model, version 1.1,'' \emph{IEEE software}, vol.~10, no.~4, pp. 18--27, 1993.

\bibitem{world2019verifiable}
W.~W.~W. Consortium \emph{et~al.}, ``Verifiable credentials data model 1.0:
  Expressing verifiable information on the web,'' \emph{https://www. w3.
  org/TR/vc-data-model/?\# core-data-model}, 2019.

\bibitem{bonawitz2017practical}
K.~Bonawitz, V.~Ivanov, B.~Kreuter, A.~Marcedone, H.~B. McMahan, S.~Patel,
  D.~Ramage, A.~Segal, and K.~Seth, ``Practical secure aggregation for
  privacy-preserving machine learning,'' in \emph{proceedings of the 2017 ACM
  SIGSAC Conference on Computer and Communications Security}, 2017, pp.
  1175--1191.

\bibitem{suzen2020novel}
A.~A. S{\"u}zen and M.~A. {\c{S}}im{\c{s}}ek, ``A novel approach to machine
  learning application to protection privacy data in healthcare: Federated
  learning,'' \emph{Nam{\i}k Kemal T{\i}p Dergisi}, vol.~8, no.~1, pp. 22--30,
  2020.

\bibitem{bennati2017primal}
S.~Bennati and C.~M. Jonker, ``Primal: A privacy-preserving machine learning
  method for event detection in distributed sensor networks,'' \emph{arXiv
  preprint arXiv:1703.07150}, 2017.

\bibitem{verachtert2021privacy}
W.~Verachtert, T.~J. Ashby, I.~Chakroun, R.~Wuyts, S.~Das, S.~Halder, and
  P.~Leray, ``Privacy preserving amalgamated machine learning for process
  control,'' in \emph{Metrology, Inspection, and Process Control for
  Semiconductor Manufacturing XXXV}, vol. 11611.\hskip 1em plus 0.5em minus
  0.4em\relax SPIE, 2021, pp. 329--341.

\bibitem{sugianto2021privacy}
N.~Sugianto, D.~Tjondronegoro, R.~Stockdale, and E.~I. Yuwono,
  ``Privacy-preserving ai-enabled video surveillance for social distancing:
  Responsible design and deployment for public spaces,'' \emph{Information
  Technology \& People}, 2021.

\bibitem{lo2021blockchain}
S.~K. Lo, Y.~Liu, Q.~Lu, C.~Wang, X.~Xu, H.-Y. Paik, and L.~Zhu,
  ``Blockchain-based trustworthy federated learning architecture,'' \emph{arXiv
  preprint arXiv:2108.06912}, 2021.

\bibitem{warnat2021swarm}
S.~Warnat-Herresthal, H.~Schultze, K.~L. Shastry, S.~Manamohan, S.~Mukherjee,
  V.~Garg, R.~Sarveswara, K.~H{\"a}ndler, P.~Pickkers, N.~A. Aziz
  \emph{et~al.}, ``Swarm learning for decentralized and confidential clinical
  machine learning,'' \emph{Nature}, vol. 594, no. 7862, pp. 265--270, 2021.

\bibitem{vassilakopoulou2020sociotechnical}
P.~Vassilakopoulou, ``Sociotechnical approach for accountability by design in
  ai systems.'' in \emph{ECIS}, 2020.

\bibitem{hax_2022}
Microsoft, ``Microsoft hax toolkit,''
  \url{https://www.microsoft.com/en-us/haxtoolkit/}, 2022, accessed: 17 Aug
  2022.

\bibitem{dai2021more}
J.~Dai, S.~Lei, L.~Dong, X.~Lin, H.~Zhang, D.~Sun, and K.~Yuan, ``More reliable
  ai solution: Breast ultrasound diagnosis using multi-ai combination,''
  \emph{arXiv preprint arXiv:2101.02639}, 2021.

\bibitem{nafreen2020architecture}
M.~Nafreen, S.~Bhattacharya, and L.~Fiondella, ``Architecture-based software
  reliability incorporating fault tolerant machine learning,'' in \emph{2020
  Annual Reliability and Maintainability Symposium (RAMS)}.\hskip 1em plus
  0.5em minus 0.4em\relax IEEE, 2020, pp. 1--6.

\bibitem{knight2002n}
J.~C. Knight, ``N-version programming,'' \emph{Encyclopedia of Software
  Engineering}, 2002.

\bibitem{lavaei2021towards}
A.~Lavaei, B.~Zhong, M.~Caccamo, and M.~Zamani, ``Towards trustworthy ai:
  safe-visor architecture for uncertified controllers in stochastic
  cyber-physical systems,'' in \emph{Proceedings of the Workshop on
  Computation-Aware Algorithmic Design for Cyber-Physical Systems}, 2021, pp.
  7--8.

\bibitem{esnaolaontology}
I.~Esnaola-Gonzalez, ``An ontology-based approach for making machine learning
  systems accountable,'' 2021.

\bibitem{shafto2012modeling}
M.~Shafto, M.~Conroy, R.~Doyle, E.~Glaessgen, C.~Kemp, J.~LeMoigne, and
  L.~Wang, ``Modeling, simulation, information technology \& processing
  roadmap,'' \emph{National Aeronautics and Space Administration}, vol.~32, no.
  2012, pp. 1--38, 2012.

\bibitem{weng2019deepchain}
J.~Weng, J.~Weng, J.~Zhang, M.~Li, Y.~Zhang, and W.~Luo, ``Deepchain: Auditable
  and privacy-preserving deep learning with blockchain-based incentive,''
  \emph{IEEE Transactions on Dependable and Secure Computing}, vol.~18, no.~5,
  pp. 2438--2455, 2019.

\bibitem{zhang2020blockchain}
W.~Zhang, Q.~Lu, Q.~Yu, Z.~Li, Y.~Liu, S.~K. Lo, S.~Chen, X.~Xu, and L.~Zhu,
  ``Blockchain-based federated learning for device failure detection in
  industrial iot,'' \emph{IEEE Internet of Things Journal}, vol.~8, no.~7, pp.
  5926--5937, 2020.

\bibitem{falco2020distributedblack}
G.~Falco and J.~E. Siegel, ``A distributedblack box'audit trail design
  specification for connected and automated vehicle data and software
  assurance,'' \emph{arXiv preprint arXiv:2002.02780}, 2020.

\bibitem{winfield2017case}
A.~F. Winfield and M.~Jirotka, ``The case for an ethical black box,'' in
  \emph{Annual Conference Towards Autonomous Robotic Systems}.\hskip 1em plus
  0.5em minus 0.4em\relax Springer, 2017, pp. 262--273.

\bibitem{falco2021governing}
G.~Falco, B.~Shneiderman, J.~Badger, R.~Carrier, A.~Dahbura, D.~Danks,
  M.~Eling, A.~Goodloe, J.~Gupta, C.~Hart \emph{et~al.}, ``Governing ai safety
  through independent audits,'' \emph{Nature Machine Intelligence}, vol.~3,
  no.~7, pp. 566--571, 2021.

\bibitem{miguel2021putting}
B.~S. Miguel, A.~Naseer, and H.~Inakoshi, ``Putting accountability of ai
  systems into practice,'' in \emph{Proceedings of the Twenty-Ninth
  International Conference on International Joint Conferences on Artificial
  Intelligence}, 2021, pp. 5276--5278.

\bibitem{oecd}
\BIBentryALTinterwordspacing
OECD, ``Tools for trustworthy ai,'' 2021. [Online]. Available:
  \url{https://www.oecd-ilibrary.org/content/paper/008232ec-en}
\BIBentrySTDinterwordspacing

\bibitem{smit2020review}
K.~Smit, M.~Zoet, and J.~van Meerten, ``A review of ai principles in
  practice,'' 2020.

\bibitem{anagnostou2022characteristics}
M.~Anagnostou, O.~Karvounidou, C.~Katritzidaki, C.~Kechagia, K.~Melidou,
  E.~Mpeza, I.~Konstantinidis, E.~Kapantai, C.~Berberidis, I.~Magnisalis
  \emph{et~al.}, ``Characteristics and challenges in the industries towards
  responsible ai: a systematic literature review,'' \emph{Ethics and
  Information Technology}, vol.~24, no.~3, pp. 1--18, 2022.

\end{thebibliography}
